\documentclass{article}

\usepackage{microtype}
\usepackage{graphicx}
\usepackage{subcaption}
\usepackage{booktabs} 

\usepackage{hyperref}
\usepackage{enumitem}

\usepackage[accepted]{icml2026}

\usepackage{amsmath}
\usepackage{amssymb}
\usepackage{mathtools}
\usepackage{amsthm}
\usepackage{algorithm}
\usepackage{algorithmic}

\usepackage[capitalize,noabbrev]{cleveref}

\theoremstyle{plain}

\theoremstyle{definition}

\theoremstyle{remark}

\usepackage[textsize=tiny]{todonotes}

\icmltitlerunning{Think-in-Control Vision-Language-Action Model for Robot Navigation in Dynamic Environments}

\begin{document}

\twocolumn[
  \icmltitle{TIC-VLA: A Think-in-Control Vision-Language-Action Model \\ for Robot Navigation in Dynamic Environments}



  \icmlsetsymbol{equal}{*}

  \begin{icmlauthorlist}
    \icmlauthor{Zhiyu Huang}{equal,ucla}
    \icmlauthor{Yun Zhang}{equal,ucla}
    \icmlauthor{Johnson Liu}{ucla}
    \icmlauthor{Rui Song}{ucla}
    \icmlauthor{Chen Tang}{ucla}
    \icmlauthor{Jiaqi Ma}{ucla}
  \end{icmlauthorlist}

  \icmlaffiliation{ucla}{University of California, Los Angeles}
  \icmlcorrespondingauthor{Zhiyu Huang}{zhiyuh@ucla.edu}
  \icmlcorrespondingauthor{Yun Zhang}{yun666@ucla.edu}
  \icmlcorrespondingauthor{Jiaqi Ma}{jiaqima@ucla.edu}

\icmlkeywords{Embodied AI, Robot Navigation, Vision-Language-Action Model}

  \vskip 0.3in
]



\printAffiliationsAndNotice{\icmlEqualContribution}

\begin{abstract}
Robots in dynamic, human-centric environments must follow language instructions while maintaining real-time reactive control. Vision-language-action (VLA) models offer a promising framework, but they assume temporally aligned reasoning and control, despite semantic inference being inherently delayed relative to real-time action.
We introduce \textbf{Think-in-Control (TIC)-VLA}, a latency-aware framework that explicitly models delayed semantic reasoning during action generation. TIC-VLA defines a delayed semantic-control interface that conditions action generation on delayed vision-language semantic states and explicit latency metadata, in addition to current observations, enabling policies to compensate for asynchronous reasoning. We further propose a latency-consistent training pipeline that injects reasoning inference delays during imitation learning and online reinforcement learning, aligning training with asynchronous deployment. To support realistic evaluation, we present \textbf{DynaNav}, a physics-accurate, photo-realistic simulation suite for language-guided navigation in dynamic environments. 
Extensive experiments in simulation and on a real robot show that TIC-VLA consistently outperforms prior VLA models while maintaining robust real-time control under multi-second reasoning latency. Project website: \href{https://ucla-mobility.github.io/TIC-VLA/}{https://ucla-mobility.github.io/TIC-VLA/}
\end{abstract}

\section{Introduction}
\label{sec:intro}

\begin{figure}[ht]
\centering
\includegraphics[width=\linewidth]{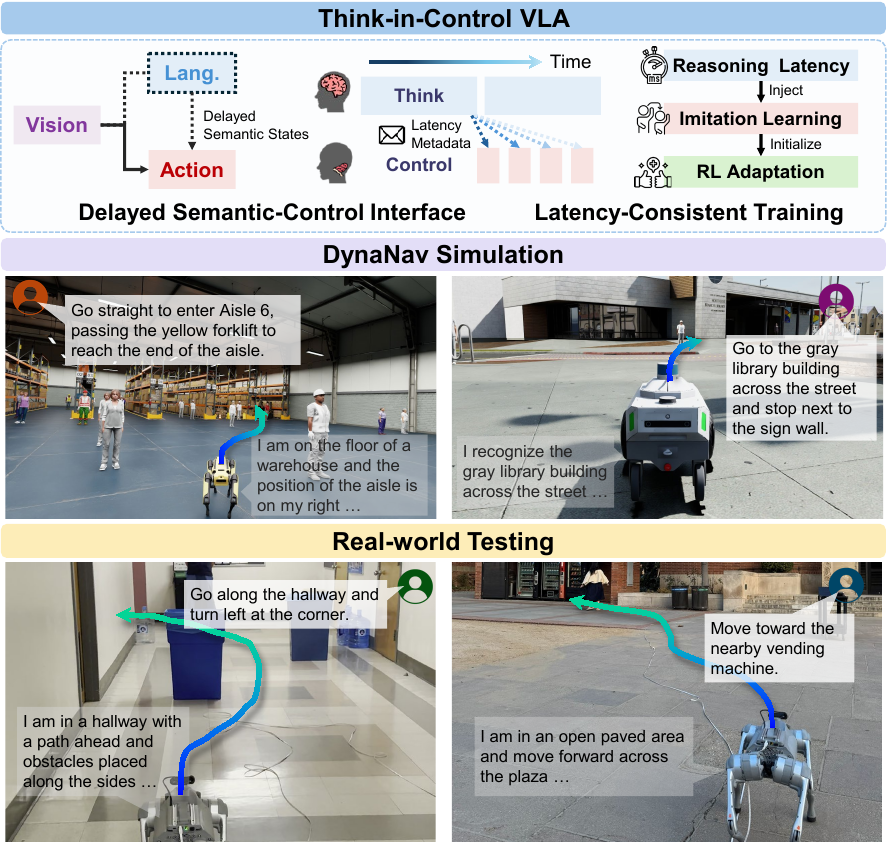}
\caption{TIC-VLA enables real-time, language-conditioned navigation by decoupling slow vision-language reasoning from fast reactive control via a delayed semantic-control interface. A latency-consistent training strategy improves robustness under variable reasoning delays. Performance is demonstrated in the DynaNav simulation and real-world indoor and outdoor navigation tasks.}
\vspace{-0.2cm}
\label{hero}
\end{figure}

Robots operating in real-world, human-centric environments must react to dynamic scenes while following high-level natural language instructions \cite{chen2025socialnav}. Vision-language-action (VLA) models \cite{hirose2025omnivla, xu2024mobility, driess2025knowledge} have emerged as a promising paradigm by unifying perception, language understanding, and control within a single learning-based system. By incorporating large vision-language models (VLMs), these approaches enable semantic grounding, task reasoning, and instruction following \cite{zhou2025autovla, choi2024embodied, zhao2025cot, lin2025onetwovla, chandaka2025human, castro2025vamos}.
However, existing VLA systems rely on a hidden and impractical assumption: reasoning and control are temporally aligned. Vision-language reasoning is produced intermittently and often with substantial delay, while control must operate continuously as the robot moves and the environment changes. As a result, semantic representations frequently correspond to past world states, yet are consumed by the policy as if they were current, introducing systematic misalignment between thinking and control.
This temporal mismatch is particularly pronounced on mobile robots with limited computation, where VLM inference can take seconds while control loops run at tens of hertz.

Most prior work on embodied navigation sidesteps this issue. Classical vision-language navigation (VLN) abstracts navigation into discrete viewpoint transitions, ignoring embodiment, dynamics, and timing \cite{krantz_vlnce_2020, yu2025correctnav, raychaudhuri2025semantic, zeng2025janusvln, eftekhar2024one, hu2025astranav}. Recent VLA systems rely on powerful computers and often pause execution during reasoning inference \cite{zhang2024navid, cheng2024navila}, which is impractical in dynamic environments. Dual-system and asynchronous VLA architectures decouple high-level reasoning from low-level control \cite{wei2025ground, InternVLA-N1_2025}, but still assume that semantic outputs are temporally fresh, implicitly treating inference latency as negligible. 
We argue that latency in reasoning is not merely an engineering inefficiency but a fundamental modeling problem. When semantic information is delayed, it may describe a past state of the robot and the environment, rather than the current one. If this delay is not represented or considered during policy learning, policies trained under idealized synchronous supervision can degrade severely when deployed under realistic inference latency.

To this end, we introduce \textbf{Think-in-Control (TIC)-VLA}, a latency-aware framework that explicitly exposes inference delay to the control policy. Rather than enforcing real-time constraints on semantic reasoning, TIC-VLA defines a \textit{delayed semantic-control interface} that allows reasoning to proceed asynchronously while enabling robust real-time control. Specifically, the reasoning module produces delayed latent semantic representations together with explicit latency and ego-motion offset metadata, while the action policy conditions on this delayed semantic-control interface. This interface is the key distinction from prior asynchronous or dual-system designs: instead of only separating slow reasoning and fast control, TIC-VLA makes the temporal staleness of the semantic state an explicit input to the controller.
Importantly, architectural decoupling alone is insufficient. Policies trained under idealized synchronous supervision fail when deployed with delayed semantic inputs. We therefore propose a \textit{latency-consistent training pipeline} in which inference delays are explicitly injected during imitation learning and reinforcement learning. This training scheme aligns the policy's training-time inputs with the delayed semantic information it receives at deployment, enabling robustness to variable reasoning latency rather than assuming synchronous semantic supervision.

To support realistic and reproducible evaluation, we develop \textbf{DynaNav}, a simulation suite featuring realistic rendering, dynamic human agents, physics-based execution, and diverse indoor and outdoor scenarios. DynaNav supports teleoperated data collection, online RL, and benchmarking. Experiments in both simulation and real-world deployment demonstrate that our proposed TIC-VLA model achieves robust navigation under realistic inference latency. \cref{hero} illustrates the key design of TIC-VLA and its performance in simulation and real-world environments. The primary contributions can be summarized as:
\begin{enumerate}[topsep=0pt, itemsep=-1.5ex]
\item We introduce TIC-VLA with a delayed semantic-control interface that provides the real-time action policy with delayed semantic features, latency metadata, and ego-motion offsets.
\item We propose a latency-consistent training pipeline that aligns learning with asynchronous inference at deployment, yielding robust navigation under variable delays.
\item We present DynaNav, a realistic simulation suite and benchmark for language-guided navigation in dynamic environments, and we demonstrate TIC-VLA's strong performance in simulation and in real-world settings.
\end{enumerate}

\begin{figure*}
    \centering
    \includegraphics[width=\linewidth]{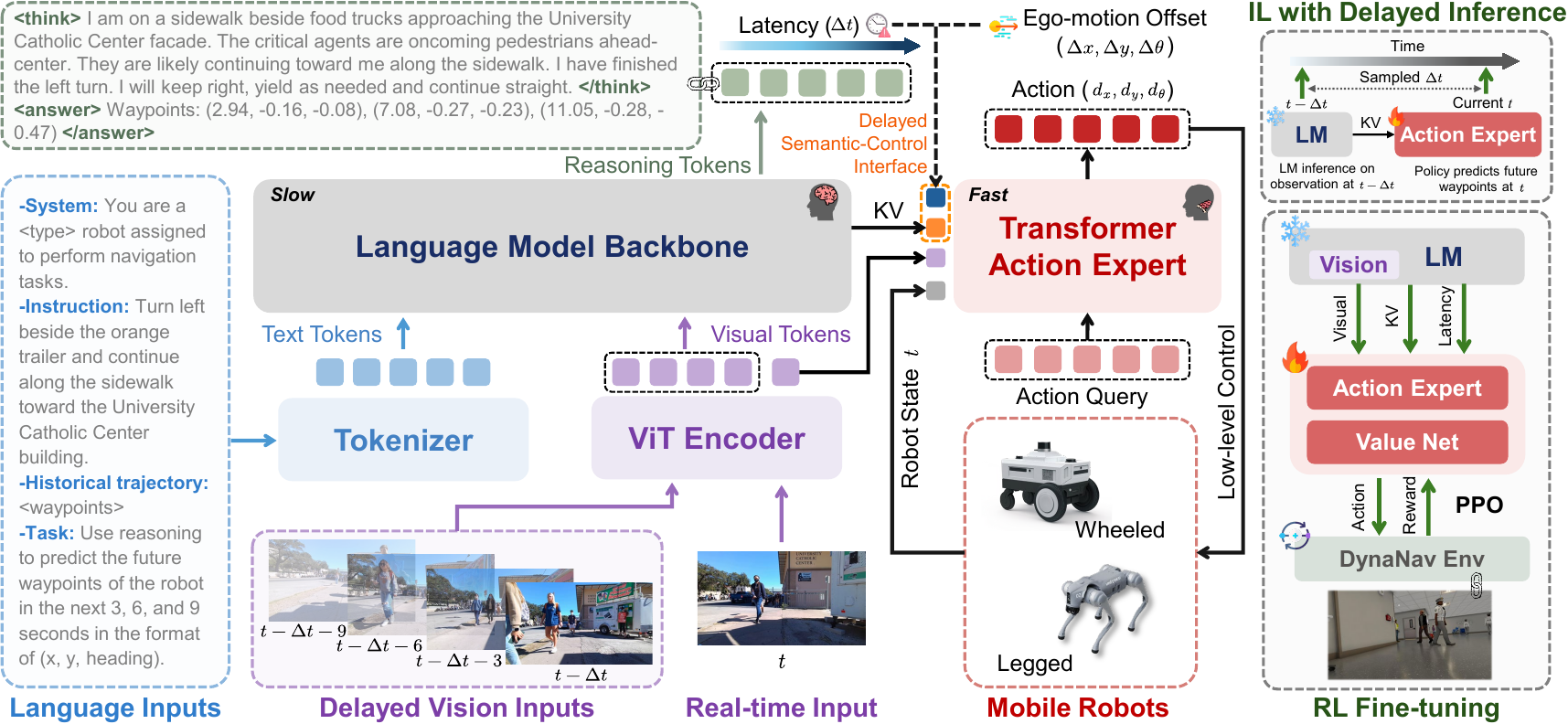}
    \caption{Overview of \textbf{TIC-VLA}. The architecture adopts a decoupled dual-system design with a fast action expert and a slow reasoning VLM. A shared vision encoder provides real-time observations to the policy and time-lagged observations to the VLM, where the delay arises naturally from slow inference. The delayed semantic-control interface (including delayed VLM KV cache features and latency metadata) is explicitly recorded. The Transformer-based action expert takes as input the current observation, robot state, and delayed semantic-control interface data to generate actions from learnable action queries via cross-attention. Multi-stage latency-consistent training combines imitation learning with delayed inference and reinforcement learning to ensure robustness to realistic conditions.}
    \label{ticvla}
    \vspace{-0.2cm}
\end{figure*}

\section{Related Work}

\textbf{Learning-based Visual Navigation}.
Recent advances in robot navigation have shifted from traditional map-based pipelines toward end-to-end learning-based models. Diffusion policies \cite{cai2025navdp, hu2025composablenav}, imitation and reinforcement learning methods \cite{liu2025compass, wu2023human, tang2025deep, he2025seeing}, and world modeling approaches \cite{liu2025x, bar2025navigation} have demonstrated strong performance in navigating complex environments without relying on maps. The integration of LLMs and VLMs into navigation tasks \cite{yuan2025opennav, xu2024mobility, zhang2025flexvln, gao2025octonav, zhou2024navgpt} has further expanded robots' semantic understanding and open-vocabulary reasoning capabilities, allowing them to follow flexible natural language commands. However, in most of these works, VLMs are employed as auxiliary modules rather than being fully integrated into the navigation pipeline. This limitation has motivated the development of VLA models, which unify perception, instruction, reasoning, and planning within a single framework.

\noindent \textbf{VLA for Navigation}.
Recent studies increasingly employ VLA models for robotic navigation \cite{du2025vl}. Representative methods span direct action prediction and intermediate planning: NaVid \cite{zhang2024navid} predicts next-step action from monocular RGB inputs, while NaVILA \cite{cheng2024navila} generates mid-level linguistic actions executed by a visual locomotion policy. TrackVLA \cite{wang2025trackvla} combines language-based recognition with diffusion-based trajectory planning. Several works further explore generalist VLA frameworks: OmniVLA \cite{hirose2025omnivla} supports multiple forms of instruction conditioning, including egocentric poses, images, and natural language, while NavFoM \cite{zhang2025embodied} demonstrates strong cross-embodiment performance. More recent systems address temporal reasoning and real-time execution, such as StreamVLN \cite{wei2025streamvln}, MobileVLA \cite{huang2025mobilevla}, and dual-system VLA approaches like DualVLN \cite{wei2025ground}, which balance deliberative reasoning and reactive control via asynchronous inference.
Despite these advances, most existing VLA-based navigation models implicitly assume negligible inference latency, often relying on powerful GPUs or blocking execution during reasoning inference. In contrast, TIC-VLA is designed for latency-aware execution and on-device deployment.

\noindent \textbf{Navigation in Dynamic Environments.}
Another line of work focuses on navigation in social and dynamic settings. Social-LLaVA \cite{payandeh2024social} fine-tunes VLMs for social navigation, while Narrate2Nav \cite{payandeh2025narrate2nav} incorporates implicit language reasoning, social cues, and human intent into visual representations. Vi-LAD \cite{elnoor2025vi} distills socially compliant navigation knowledge from large VLMs into lightweight policies for real-time deployment. However, these methods typically rely on specialist policies at inference time and do not keep an in-the-loop VLM during execution.
Evaluation is further limited by existing simulators. Arena \cite{shcherbyna2025arena} and UrbanSim \cite{wu2025towards} support dynamic agents but lack language-conditioned navigation. SocialNav-SUB \cite{munje2025socialnav} evaluates real-world social navigation without a controllable simulator, while HA-VLN \cite{dong2025ha} relies on simplified observations and abstracted control. In contrast, \textit{DynaNav} provides a physics-accurate simulation suite for language-guided navigation in human-centric environments.

\section{Method}

\subsection{Problem Formulation}
We consider an instruction-following navigation problem in which an embodied agent must follow natural language instructions to navigate dynamic environments. At each control timestep \( t \), the agent receives: (1) a natural language instruction and context \( \mathcal{I} \), specifying the navigation goal and historical trajectory; (2) an egocentric observation history \( \mathcal{O}_t = \{x_0, \ldots, x_t\} \), consisting of RGB frames \( x_t \in \mathbb{R}^{H \times W \times 3} \); and (3) the robot state \( s_t \in \mathbb{R}^3 \), encoding ego-motion information such as linear velocity and angular velocity. Conditioned on these inputs, the agent must output an action $\mathbf{a}_t$ at each timestep to safely and efficiently progress toward the goal. The environment evolves in response to the agent’s actions, and the episode continues until the agent reaches the goal or terminates.

We consider settings where a large-scale VLM is employed to interpret the natural-language instructions and provide semantic guidance. While such models are powerful, their reasoning often introduces non-negligible inference latency $\Delta t$. As a result, semantic outputs may become temporally misaligned with the agent’s current observations and state, creating a key challenge for real-time navigation.


\subsection{Think-in-Control VLA}
\label{subsec:ticvla}

An overview of the TIC-VLA framework is shown in \cref{ticvla}. TIC-VLA adopts a dual-system execution paradigm in which high-level semantic reasoning and low-level control operate asynchronously. Crucially, rather than treating this as an architectural contribution, we explicitly model the resulting inference delay as part of the control problem. The key design principle is to expose reasoning latency to the action policy and train the policy to act under delayed semantic observations. We employ InternVL3-1B \cite{zhu2025internvl3} as the vision-language backbone for semantic and instruction understanding.
At a high level, a VLM performs semantic reasoning over delayed visual context and language instructions, while a reactive action policy executes at a high control frequency and never blocks on VLM inference. The action policy conditions on cached semantic representations together with explicit latency and ego-motion metadata, allowing it to interpret the delayed semantic information in the correct temporal context. This latency-aware semantic-control coupling enables robust navigation despite asynchronous and delayed reasoning updates.

\textbf{VLM Semantic Reasoning.}
We formalize inference latency as a core variable in the system. We define the \emph{effective reasoning latency} as \( \Delta t = t_\text{infer} +  t_\text{elapse} \ge 0 \), which accounts for both the VLM inference time ($t_\text{infer}$) and the elapsed time since the last completed reasoning update ($t_\text{elapse}$). At the current timestep \( t \), the VLM operates on visual observations anchored at time \( t - \Delta t \), rather than the current frame.
Given a set of historical frames
$
\mathcal{X}^{\text{vlm}}_{t-\Delta t}
=
\{x_{t-\Delta t-\delta} \mid \delta \in \{0, 3, 6, 9\}\}
$,
the VLM performs semantic reasoning conditioned on the instruction \( \mathcal{I} \). The resulting output, denoted \( \mathcal{R}_{t-\Delta t} \), encodes high-level scene understanding, critical object identification, intent prediction, and future target waypoints derived from delayed observations. The reasoning results, including predicted waypoints, are generated relative to the time of inference start \( t-\Delta t \), rather than the current control timestep.
This explicit temporal anchoring allows the downstream policy to know when the reasoning was produced, rather than treating the reasoning results as instantaneous or noisy signals.

\textbf{Latency-Aware Action Policy}.
The action policy \( \pi_\theta \) runs at a high frequency and is responsible for real-time planning. At each timestep \( t \), it conditions on four categories of inputs: (1) the current visual observation \( x_t \); (2) the current robot state \( s_t \); (3) the most recent semantic hidden state \( \mathcal{S}_{t-\Delta t} \) produced by the VLM (i.e., the last-layer key-value cache); (4) explicit latency metadata, including the effective latency \( \Delta t \) and the corresponding motion offsets \( \Delta \mathbf{p}_t = (\Delta x, \Delta y, \Delta \theta) \) accumulated since reasoning generation.
Providing both delayed semantic states and latency metadata establishes a \textit{delayed semantic-control interface}: semantic features describe a past state of the world, while the control policy is responsible for reinterpreting them in the current frame. This allows the policy to reason consistently about delayed semantics as the robot moves during inference.
The policy outputs a short horizon of future actions:
\begin{equation}
\mathbf{a}_t = \{a_t^{1}, \dots, a_t^{T}\}
= \pi_{\theta}\big(\mathcal{S}_{t-\Delta t}, x_t, s_t, \Delta t, \Delta \mathbf{p}_t \big),
\end{equation}
where each \( a_t^{i} \in \mathbb{R}^3 \) represents a continuous action. The action chunks are integrated into a short-horizon trajectory, and a target point is chosen for execution.

As shown in \cref{fig3}(a), the action policy utilizes a dedicated action query token that attends to the scene context through a stack of cross-attention Transformer layers. Visual tokens and VLM cache features are first projected into a shared latent space via MLP layers, while the robot state and latency metadata are encoded and added with positional embeddings. These inputs are concatenated as key-value tokens, and the updated action query representation is passed through an MLP to generate the action outputs.

\textbf{Asynchronous Semantic Reasoning and Control.} 
TIC-VLA operates in a closed-loop asynchronous manner. The VLM periodically updates semantic reasoning based on delayed visual inputs, while the action policy continuously executes without waiting for inference to complete. The cached VLM hidden state is updated as follows:
\begin{equation}
\small
(\mathcal{S}^{\text{cache}}_t, \mathcal{R}^{\text{cache}}_t) =
\begin{cases}
(\mathcal{S}_t, \mathcal{R}_t), \text{if inference finishes at } t, \\
(\mathcal{S}^{\text{cache}}_{t^-}, \mathcal{R}^{\text{cache}}_{t^-}), \text{otherwise},
\end{cases}
\end{equation}
where $t^-$ denotes the most recent timestep prior to $t$ at which inference was completed.
At every control timestep, the action policy conditions on the most recent cached VLM hidden state
$\mathcal{S}_{t-\Delta t} = \mathcal{S}^{\text{cache}}_t$. In addition, by conditioning action generation on latency metadata and ego-motion, TIC-VLA can reinterpret stale semantic information in the current state. An illustration of the asynchronous inference process is shown in \cref{fig3}(d), and latency is measured as the sum of two parts: VLM reasoning inference time and the elapsed time since the last finished inference.

\subsection{Latency-Consistent Training Pipeline}
\label{sec:training}

\begin{figure}[ht]
    \centering
    \includegraphics[width=\linewidth]{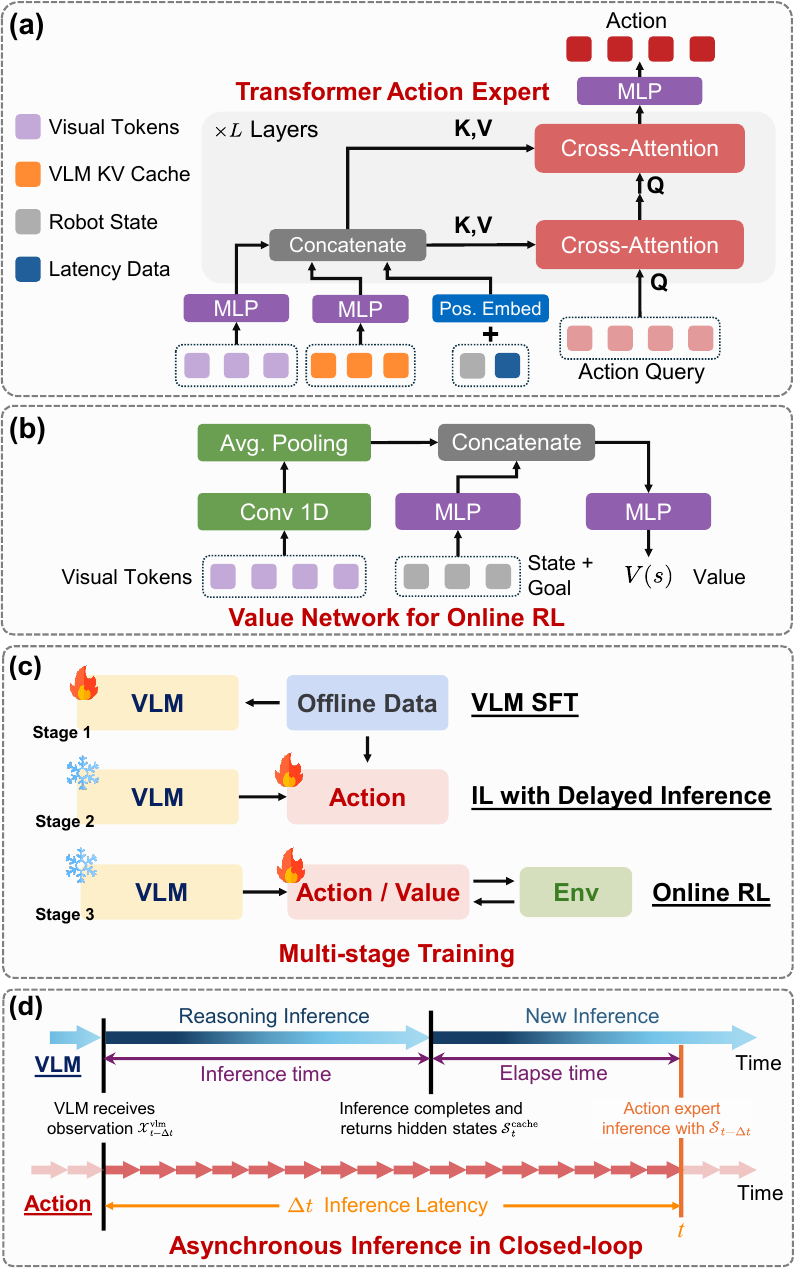}
    \caption{
    Details of TIC-VLA action policy structure, training, and asynchronous execution.
    (a) Latency-aware action policy that predicts action chunks from multimodal inputs.
    (b) Value network used during online reinforcement learning.
    (c) Three-stage latency-consistent training pipeline combining VLM supervision, imitation learning, and reinforcement learning.
    (d) Asynchronous inference and control with explicit latency modeling.
    }
    \label{fig3}
    \vspace{-0.4cm}
\end{figure}

We adopt a three-stage training pipeline designed to enforce \emph{latency consistency} between training and deployment. An overview of the pipeline is provided in \cref{fig3}(c).

\textbf{Supervised Fine-tuning of the VLM.}
We first fine-tune the VLM on structured semantic reasoning data collected from both simulation and real-world environments. Training samples are automatically annotated using GPT-5. Given past and future image sequences, the robot’s positional context, and the corresponding trajectory, GPT-5 produces: (1) a long-horizon instruction describing the navigation goal, and (2) a concise, structured reasoning trace capturing critical object identification, intention prediction, and resulting action. Details are provided in the supplementary material.

During this stage, the vision encoder is kept frozen. The language model is trained to produce reasoning tokens and waypoint predictions conditioned on visual tokens and instructions. We optimize the standard autoregressive cross-entropy loss over the target token sequence:
\begin{equation}
\mathcal{L}_l = - \frac{1}{N_l} \sum_{t=1}^{N_l} \log p_{\phi}(y_t \mid y_{<t}, \mathcal{I}, \mathbf{V}),
\end{equation}
where $y_t$ denotes the $t$-th token in the target sequence, $N_l$ is the sequence length, and $\phi$ are the language model parameters. 
We mix waypoint-only and scene-reasoning-augmented targets during training for flexible prompting at inference time.

\textbf{Imitation Learning under Reasoning Latency.}
To compensate for the uncertain delay in semantic reasoning, we perturb originally aligned and synchronous demonstrations to synthesize training data with delayed semantic reasoning. Specifically, we sample reasoning delays $\Delta t$ uniformly from \( [0, 10] \) seconds and condition the policy on: (1) the current image input and robot state, (2) KV cache features from the delayed VLM reasoning output, (3) explicit latency metadata. This exposes the policy to a range of temporally misaligned semantic representations during training.
The action policy $\pi_{\theta}$ is trained via imitation learning using human demonstration trajectories. As low-level control actions are not available, we integrate the predicted actions forward to obtain positions $(x, y, \theta)$ over the prediction horizon and compare them against ground-truth trajectories using a smooth $L_1$ loss:
\begin{equation}
\mathcal{L}_a = \frac{1}{T} \sum_{i=1}^{T} \text{SmoothL1}\big(\hat{p}_t^{(i)} - {p}_t^{(i)}\big),
\end{equation}
where $\hat{p}_t^{(i)}$ is the predicted pose at sub-timestep $i$, and ${p}_t^{(i)}$ is the ground-truth pose.

\textbf{Reinforcement Learning with Asynchronous Guidance.}
While imitation learning with synthesized reasoning delays provides a strong initialization that accounts for inference latency, the resulting training data distribution remains mismatched to the closed-loop distribution induced by coupled, asynchronous reasoning-action interactions. Motivated by prior work \cite{lu2025vla, li2025simplevla, chen2025era}, we fine-tune only the action policy using RL while keeping the vision encoder and language model frozen. This allows the policy to learn to interpret delayed VLM guidance, handle dynamic agents, and mitigate variability introduced by asynchronous inference.

We construct a simulation environment with dynamic human agents and train the policy using Proximal Policy Optimization (PPO) \cite{schulman2017proximal}. The value network, shown in \cref{fig3}(b), takes as input the current image tokens, the goal position, and the robot state, and outputs the estimated state value. The policy outputs a Gaussian action distribution, where the action derived from the predicted trajectory is the mean, and the standard deviation is learned during training. The reward function is defined as:
\begin{equation}
\label{reward}
r_t = w_g r^{\text{goal}}_t + w_p r^{\text{progress}}_t + w_c r^{\text{collision}}_t + w_s r^{\text{speed}}_t,
\end{equation}
where $r^{\text{goal}}_t$ rewards reaching the target, $r^{\text{progress}}_t$ encourages progress toward the goal, $r^{\text{collision}}_t$ penalizes collisions with humans or static obstacles, and $r^{\text{speed}}_t$ penalizes both excessively slow motion and excessive speed. $w_g, w_p, w_c,$ and $w_s$ are weights for these terms.

To further improve robustness under asynchronous deployment, we inject stochastic inference delays following each VLM update to mimic the latency characteristics observed on edge hardware. This enforces consistency between training and execution conditions. Additional implementation details are provided in the supplementary material.

\subsection{DynaNav}

Evaluating our proposed latency-aware VLA framework requires more realistic navigation benchmarks than classic VLN benchmarks such as R2R \cite{li2019robust}, VLN-CE \cite{krantz_vlnce_2020}, and RxR \cite{ku2020room}, which operate in small indoor environments and abstract navigation as viewpoint transitions without physical interaction. VLN-PE \cite{wang2025rethinking} and GRUtopia \cite{grutopia} support embodied evaluation but do not model navigation among dynamic human participants, while SocialHM3D \cite{gong2024cognition} and HA-VLN \cite{dong2025ha} incorporate human agents with limited visual realism. To fill this gap, DynaNav provides a realistic benchmark integrating language-guided navigation, large-scale scenes, diverse human agents, and physics-based robot control.

\textbf{Simulation Environments.}
We construct simulation environments in Isaac Sim \cite{IsaacSim2025}, with realistic, controllable dynamic interactions. Four representative scenes (i.e., warehouse, hospital, office, and outdoor sidewalk) are designed to capture a broad range of navigation contexts. Human behaviors are modeled using Isaac Sim’s built-in human simulation tools, supporting behaviorally plausible pedestrian movement. Our simulation setup supports both wheeled (Nova Carter) and quadruped (Boston Dynamics Spot) robots. We develop custom robot behavior scripts that allow two modes of operation: (1) Human teleoperation mode, which enables manual control for collecting expert demonstrations, and (2) End-to-end model control mode, which allows direct control of the robot with end-to-end planning models.

\textbf{Designed for Scalable RL Training.}
We employ Isaac Lab \cite{mittal2025isaaclab} and leverage its GPU-accelerated simulation to build environments for scalable reinforcement learning training. We implement a custom human behavior control script to generate human movements within the environment. Human-robot and robot-scene interactions are fully physics-based with realistic contact dynamics. This setup allows us to run multiple parallel environments, enabling scalable and efficient RL training.

\textbf{Diverse Benchmark Tasks.}
We develop a comprehensive benchmark of 85 test cases to evaluate navigation performance across diverse conditions. Tasks vary along three dimensions: (1) \textit{Crowd Density}, ranging from empty spaces to densely populated settings, capturing different levels of dynamic complexity. (2) \textit{Navigation Distance}, adjusted from short-term navigation to long-horizon planning, reflecting increasing navigation difficulty. (3) \textit{Scene Type}, with evaluation conducted across four distinct environments to assess robustness to varying spatial layouts and human behaviors.
For each task, standardized initial and goal positions are specified with a language instruction. Additional details are provided in the supplementary material.

\begin{figure*}
    \centering
    \includegraphics[width=\linewidth]{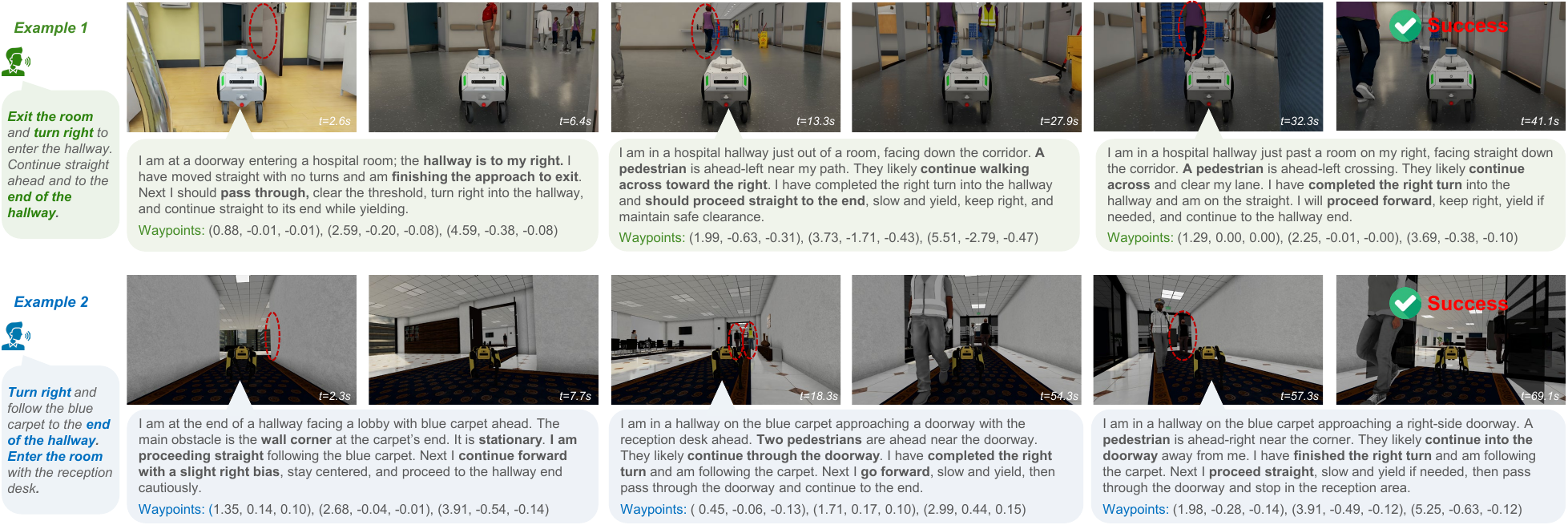}
    \caption{Qualitative closed-loop results of TIC-VLA in DynaNav hospital (top) and office (bottom) environments. TIC-VLA demonstrates effective semantic reasoning while producing reactive navigation actions in dynamic scenarios.}
    \label{fig:sim_dynanav}
    \vspace{-0.3cm}
\end{figure*}

\section{Experiments}
\subsection{Experimental Setup}

\textbf{Datasets}.
We train the model using three datasets featuring dynamic human-robot interactions: (1) SCAND \cite{karnan2022socially}, which contains 8.7 hours of robot-driven trajectories across diverse social environments; (2) GND \cite{liang2025gnd}, which comprises over 11 hours of recorded data collected in various campus environments; (3) DynaNav simulation dataset, collected using our designed dynamic simulation environments, containing 5.1 hours of robot navigation data across multiple scene types.

\textbf{Implementation Details}.
For VLM SFT, we employ full-parameter training due to the compact size of TIC-VLA. Training is performed using Distributed Data Parallel on eight NVIDIA L40S GPUs, with a batch size of 2 per GPU. AdamW is used as the optimizer with a cosine learning rate schedule, initialized at \(2\times10^{-5}\). For training the action expert, we increase the batch size to 16 per GPU and set the initial learning rate to \(2\times10^{-4}\). RL fine-tuning of the action policy is conducted on a single NVIDIA L40S GPU for 400 iterations across three tasks in three environments. Additional details, including hyperparameters and data preprocessing, are provided in the supplementary material.

\textbf{Baselines.}
We evaluate TIC-VLA against two categories of methods. Point-goal navigation policies are included as reference baselines to contextualize task difficulty:
(1) a vanilla Behavior Cloning (BC) policy that maps RGB observations and point-goal commands directly to actions;
(2) a vanilla RL policy trained on RGB observations and point-goal commands;
(3) NavDP \cite{cai2025navdp}, a point-goal-conditioned diffusion-based navigation policy. The primary language-guided navigation baselines are listed below:
(1) NaVILA \cite{cheng2024navila}, a hierarchical VLA model that translates language instructions into mid-level commands;
(2) Uni-NaVid \cite{zhang2024uni}, a unified video-based VLA model trained across multiple navigation tasks; 
(3) DualVLN~\cite{wei2025ground}, a dual-system VLA model for language-guided navigation;
(4) MobileVLA-R1~\cite{huang2025mobilevla}, a reasoning-enhanced VLA framework for mobile robots with chain-of-thought supervision; 
(5) OmniVLA~\cite{hirose2025omnivla}, an omni-modal VLA navigation model that supports language, visual-goal, and pose-goal conditioning.
All language-guided baselines are fine-tuned on the same training datasets and evaluated under the same settings for a fair comparison.

\noindent\textbf{Evaluation metrics}. 
We adopt the following evaluation metrics: 
(1) Navigation Error (NE): the final distance between the agent and the goal;
(2) Success Rate (SR): the percentage of episodes in which the agent stops within 1.5 meters of the goal;
(3) Success weighted by Path Length (SPL): SR weighted by the ratio between the shortest path length and the actual path length, penalizing inefficient trajectories; 
(4) Collision Rate (CR): the percentage of episodes in which the agent collides with static obstacles or humans, quantifying the safety of navigation behavior.

\textbf{Real-world experimental setup}.
To evaluate real-time performance on edge devices, we deploy the model on two platforms with different power budgets: an NVIDIA Jetson Orin NX (25W) and an RTX 4060 Laptop GPU (50W), representing typical edge computing capabilities. An RTX A6000 GPU is used only when the baselines cannot run on these devices. The navigation policy is executed on a Unitree Go2 quadruped robot in real-world navigation tasks. We employ FlashAttention to improve inference efficiency. Performance is measured by the average success rate. An episode is considered a failure if manual intervention is required to prevent collisions.

\subsection{Simulation Testing}
\textbf{Performance on the DynaNav benchmark.}
All experiments are conducted on an NVIDIA L40S GPU, with the action policy running at 10~Hz and asynchronous VLM reasoning running at 0.5~Hz. Results are summarized in \cref{tab:main_results}. Point-goal methods bypass vision-language inference and therefore incur no reasoning latency, but they also rely on privileged goal information. In contrast, TIC-VLA uses only egocentric observations and language instructions, without access to privileged goals or maps.
TIC-VLA achieves strong performance under this setting. Without RL fine-tuning, TIC-VLA is competitive with NavDP, a point-goal method with privileged state access, and outperforms the vanilla BC and RL baselines. After RL fine-tuning, TIC-VLA achieves the highest success rate and the lowest collision rate, indicating improved closed-loop robustness in dynamic scenes. Although NavDP obtains the lowest navigation error and the highest SPL, it benefits from direct point-goal supervision.
TIC-VLA also clearly outperforms prior VLN baselines, including Uni-NaVid, NaVILA, and DualVLN. Their lower success rates and higher collision rates suggest that methods designed for more abstract navigation settings struggle with continuous control and dynamic obstacles. Moreover, TIC-VLA outperforms stronger VLA baselines, including OmniVLA and MobileVLA. The synchronous TIC-VLA variant also degrades substantially, confirming that blocking control on slow VLM inference harms real-time navigation. \cref{fig:sim_dynanav} shows qualitative examples of TIC-VLA's semantic reasoning and reactive navigation in dynamic environments. We provide additional dynamic video results on the \href{https://ucla-mobility.github.io/TIC-VLA/#results}{project website}.

\begin{table}[ht]
\centering
\caption{Performance of TIC-VLA and baseline methods on the DynaNav benchmark. BC, RL, and NavDP are point-goal-based.}
\vspace{-0.2cm}
\label{tab:main_results}
\renewcommand{\arraystretch}{1.0}
\resizebox{\linewidth}{!}{
\begin{tabular}{l|cccc}
\toprule
\textbf{Method} & \textbf{NE ($\downarrow$)} & \textbf{SR ($\uparrow$)} & \textbf{SPL ($\uparrow$)} & \textbf{CR ($\downarrow$)} \\
\midrule
BC Policy               & 9.96  & 45.88 & 41.52 & 35.29 \\
RL Policy               & 12.20 & 30.59 & 28.45 & 36.47 \\
NavDP                   & \textbf{8.61} & 54.12 & \textbf{52.62} & 30.59 \\ 
\midrule
Uni-NaVid               & 15.90 & 22.35 & 19.61 & 49.41 \\
NaVILA                  & 17.20 & 28.24 & 25.51 & 48.24 \\
DualVLN                 & 16.45 & 30.59 & 27.82 & 47.06 \\ 
OmniVLA                 & 16.53 & 31.76 & 28.33 & 49.41 \\ 
MobileVLA               & 16.09 & 32.94 & 30.31 & 45.88 \\
\midrule
\textbf{TIC-VLA (Sync.)}  & 16.31 & 32.94 & 29.64 & 41.18 \\
\textbf{TIC-VLA (no RL)} & 10.85 & 47.06 & 42.41 & 34.12 \\
\textbf{TIC-VLA}         & 10.55 & \textbf{55.29} & 50.29 & \textbf{28.24} \\
\bottomrule
\end{tabular}
}
\vspace{-0.2cm}
\end{table}

\begin{figure}[ht]
    \centering
    \includegraphics[width=\linewidth]{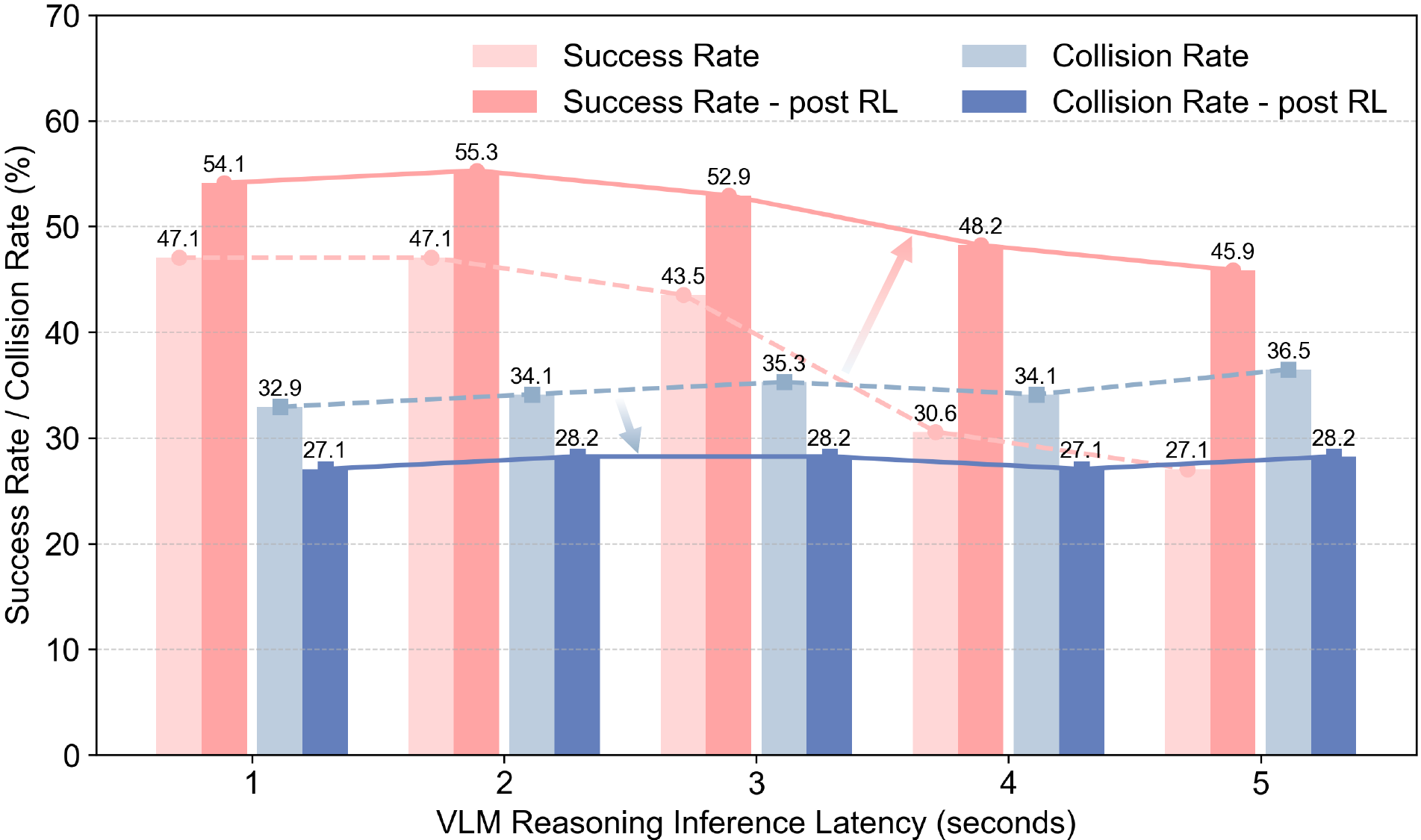}
    \caption{
    The effect of VLM asynchronous reasoning inference latency in TIC-VLA on task performance.
    }
    \label{fig:inferencetime}
    \vspace{-0.2cm}
\end{figure}

\begin{figure*}[t]
    \centering
    \includegraphics[width=0.99\linewidth]{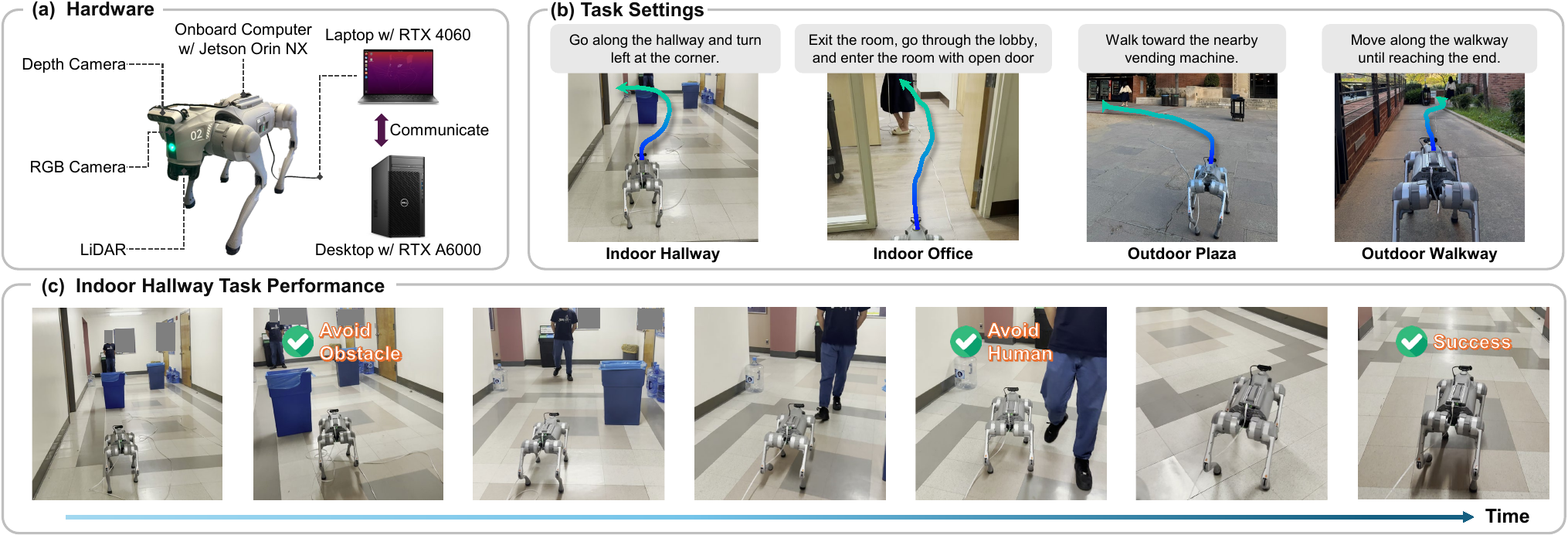}
    \caption{Real-world evaluation of TIC-VLA. (a) Hardware configuration, including the robot platform and computation setup. (b) Designed indoor and outdoor vision-language navigation tasks. (c) Qualitative results from an indoor hallway navigation task, showing the robot following natural language instructions while avoiding obstacles and humans and reaching the goal.}
    \label{fig:test}
    \vspace{-0.2cm}
\end{figure*}

\textbf{Latency Robustness Analysis.}
\cref{fig:inferencetime} reports performance under increasing VLM reasoning inference latency for TIC-VLA in simulation, before and after RL fine-tuning. As latency increases, the IL-based action expert exhibits a noticeable decline in success rate, indicating higher sensitivity to delayed reasoning updates. In contrast, the RL-fine-tuned policy maintains consistently higher success rates across all latency settings, demonstrating improved robustness to inference latency. This result highlights the effectiveness of RL fine-tuning in mitigating latency-induced performance degradation. Collision rates are relatively insensitive to inference latency, suggesting that the asynchronous policy preserves reactivity independent of reasoning speed.

\textbf{Influence of Semantic-Control Interface and Latency Awareness.}
We evaluate the impact of different delayed semantic-control interfaces and latency-awareness on TIC-VLA. Specifically, we compare interface variants that use waypoint-based guidance and KV-cache-based features, each trained with and without explicit latency-aware modeling and training. As shown in \cref{tab:guidance_delay}, using KV-cache features significantly improves navigation success, and latency-awareness enhances performance under asynchronous inference. The waypoint-based interface leads to inferior performance due to its sparsity and potential inconsistency with the agent’s local observations. In contrast, KV-cache features preserve richer semantic and contextual information from the VLM, allowing the control policy to better adapt to delayed reasoning outputs. Combining both the KV cache feature interface and latency-aware modeling and training achieves the best overall performance.

\begin{table}[ht]
\centering
\caption{Influence of semantic interface and latency training.}
\vspace{-0.2cm}
\label{tab:guidance_delay}
\renewcommand{\arraystretch}{1.0}
\resizebox{\linewidth}{!}{
\begin{tabular}{lc|cccc}
\toprule
\textbf{Interface} & \textbf{Latency} & \textbf{NE ($\downarrow$)} & \textbf{SR ($\uparrow$)} & \textbf{SPL ($\uparrow$)} & \textbf{CR ($\downarrow$)} \\
\midrule
Waypoint & \texttimes & 21.17           & 16.47         & 15.89             & 47.06 \\
Waypoint & \checkmark & 20.32           & 22.35         & 18.34             & 42.35 \\
KV Cache & \texttimes & 16.74           & 30.59         & 28.31             & 40.00\\
KV Cache & \checkmark & \textbf{10.85}  & \textbf{47.06}& \textbf{42.41}    & \textbf{34.12}\\
\bottomrule
\end{tabular}
}
\vspace{-0.2cm}
\end{table}

\subsection{Real-world Testing}
We evaluate TIC-VLA on a Unitree Go2 robot across four real-world navigation tasks: (1) an indoor hallway with dynamic human and static obstacles, (2) a cluttered office, (3) an outdoor plaza, and (4) an outdoor walkway with uneven terrain. These scenarios cover indoor and outdoor deployment conditions with challenges such as moving pedestrians, narrow passages, clutter, and terrain variation. Task descriptions and hardware configurations are shown in \cref{fig:test}. All evaluations are conducted zero-shot, without task-specific training data. For each task, we conduct five trials and report the average success rate. During deployment, RGB images from the front-facing camera are streamed to TIC-VLA, where VLM reasoning runs asynchronously from the high-frequency action policy.

\begin{table}[ht]
\centering
\caption{Real-world testing results. Runtimes for dual-system methods are reported as action policy / VLM reasoning latency.}
\vspace{-0.2cm}
\label{tab:method_platform}
\renewcommand{\arraystretch}{1.0}
\resizebox{\linewidth}{!}{
\begin{tabular}{lc|ccc}
\toprule
\textbf{Method} & \textbf{Platform} & \textbf{Success Rate ($\uparrow$)} & \textbf{Runtime (ms)} \\
\midrule
TIC-VLA (no RL) & 4060 & 0.70 & -- \\
TIC-VLA & 4060 & \textbf{0.85} & 85.73 / 3430.73 \\
TIC-VLA & Orin NX & 0.75 & 120.27 / 4831.73 \\
TIC-VLA & A6000 & 0.80 & 32.70 / 1681.66 \\
Dual-VLN (7B) & A6000 & 0.50 & 299.92 / 1534.67 \\
NaVILA (7B) & A6000 & 0.35 & 4106.62 \\
\bottomrule
\end{tabular}
}
\vspace{-0.2cm}
\end{table}

We compare TIC-VLA with DualVLN~\cite{wei2025ground} and NaVILA~\cite{cheng2024navila} under the same real-world settings. As shown in \cref{tab:method_platform}, TIC-VLA achieves the highest success rate with a smaller model and lower control-side latency. RL fine-tuning further improves success from 0.70 to 0.85, indicating stronger robustness in physical deployment, especially around pedestrians and obstacles.
Across compute platforms, TIC-VLA maintains a 0.75 success rate on the Jetson Orin NX despite multi-second VLM reasoning latency, demonstrating its suitability for edge deployment. Although the A6000 reduces VLM latency, its success rate does not exceed local RTX 4060 deployment, likely due to remote communication overhead. These results show that real-world performance depends not only on inference speed, but also on communication latency and closed-loop control frequency.
Overall, TIC-VLA transfers zero-shot to real-world robot navigation and remains robust under asynchronous high-latency VLM inference, demonstrating the practical value of explicit latency modeling.

\subsection{Ablation Study}

\textbf{Influence of Reasoning at Test Time.}
We evaluate explicit VLM reasoning during inference by comparing the RL fine-tuned model with and without reasoning-token outputs. As shown in \cref{tab:reasoning_ablation}, enabling reasoning clearly improves navigation performance, reducing NE from 14.23 to 10.55 while increasing SR from 40.00 to 55.29 and SPL from 34.22 to 50.29. Disabling reasoning reduces VLM overhead and increases the forward rate from 0.5~Hz to 4~Hz, but leads to weaker goal completion and navigation progress. Although the non-reasoning variant has a lower collision rate, this mainly reflects reduced activity and more frequent failure rather than safer navigation. In contrast, the reasoning-enabled model interacts more actively with the environment while maintaining substantially higher task completion and navigation efficiency. These results indicate that test-time reasoning improves task success, while TIC-VLA mitigates its latency through explicit latency-aware control.

\begin{table}[ht]
\centering
\small
\caption{Effect of VLM reasoning at test time.}
\vspace{-0.2cm}
\renewcommand{\arraystretch}{1.05}
\begin{tabular}{l|cccc}
\toprule
\textbf{Inference}          & \textbf{NE ($\downarrow$)} & \textbf{SR ($\uparrow$)} & \textbf{SPL ($\uparrow$)} & \textbf{CR ($\downarrow$)}   \\
\midrule
W/o Reasoning               & 14.23             & 40.00          & 34.22          & \textbf{25.88} \\
W/ Reasoning                & \textbf{10.55}    & \textbf{55.29} & \textbf{50.29} & 28.24 \\
\bottomrule
\end{tabular}
\label{tab:reasoning_ablation}
\vspace{-0.2cm}
\end{table}

\textbf{Influence of Action Prediction Horizon.}
We ablate the action prediction horizon to study the trade-off between short-term responsiveness and longer-horizon planning. A shorter horizon makes the policy more reactive but provides less planning context, whereas a longer horizon provides more temporal context but can make near-term action prediction less accurate.
As shown in \cref{tab:horizon_ablation}, the 3-second horizon achieves the best overall performance among TIC-VLA variants without RL fine-tuning. The 1-second horizon obtains the lowest collision rate, suggesting stronger short-term reactivity, but yields a lower success rate and SPL. The 5-second horizon also underperforms the 3-second setting, indicating that overly long prediction horizons may reduce control accuracy. We therefore use a 3-second prediction horizon as the default setting.

\begin{table}[ht]
\centering
\small
\caption{Effect of action prediction horizon. Results are reported without RL fine-tuning.}
\vspace{-0.2cm}
\renewcommand{\arraystretch}{1.05}
\begin{tabular}{l|cccc}
\toprule
\textbf{Horizon} & \textbf{NE ($\downarrow$)} & \textbf{SR ($\uparrow$)} & \textbf{SPL ($\uparrow$)} & \textbf{CR ($\downarrow$)} \\
\midrule
1s & 12.23 & 42.35 & 38.72 & \textbf{32.94} \\
3s (ours) & \textbf{10.85} & \textbf{47.06} & \textbf{42.41} & 34.12 \\
5s & 11.77 & 40.00 & 36.46 & 35.29 \\
\bottomrule
\end{tabular}
\label{tab:horizon_ablation}
\vspace{-0.2cm}
\end{table}

\textbf{Influence of Ego-motion Offset in Action Policy.}
\cref{tab:offset} shows the effect of incorporating ego-motion offset into the latency-aware action policy. Without ego-motion offset, the policy receives delayed semantic guidance but lacks explicit information about how far the robot has moved since the guidance was generated. As a result, stale VLM features may be treated as temporally aligned with the current observation, leading to degraded navigation accuracy, lower success rates, and less efficient trajectories.
By explicitly modeling the ego-motion accumulated during the inference delay, the policy can reinterpret delayed semantic features in the current robot frame. This improves the alignment between semantic reasoning and real-time control, allowing the action expert to compensate for robot displacement and heading changes during asynchronous VLM inference, resulting in substantially improved performance.

\begin{table}[ht]
    \centering
    \small
    \caption{Effect of incorporating ego-motion offset into the latency-aware action policy. Results are reported without RL fine-tuning.}
    \vspace{-0.2cm}
    \renewcommand{\arraystretch}{1.05}
    \label{tab:offset}
    \begin{tabular}{l|cccc}
    \toprule
    \textbf{Method} & \textbf{NE ($\downarrow$)} & \textbf{SR ($\uparrow$)} & \textbf{SPL ($\uparrow$)} & \textbf{CR ($\downarrow$)} \\
    \midrule
    W/o Motion Offset    &     12.97   & 41.18 &  36.36  &  36.47   \\
    W/ Motion Offset    &  \textbf{10.85}  & \textbf{47.06}& \textbf{42.41} & \textbf{34.12} \\
    \bottomrule
    \end{tabular}
    \vspace{-0.4cm}
\end{table}

\section{Conclusions}
We introduce TIC-VLA, a latency-aware VLA framework that addresses the temporal mismatch between slow semantic reasoning and real-time control. By using a delayed semantic-control interface and training policies under realistic inference delays, TIC-VLA enables robust language-guided navigation under substantial latency. Simulation and real-world results show consistent improvements over prior navigation and language-guided baselines.

\textbf{Limitations.}
TIC-VLA has three main limitations. First, the current system is not yet fully optimized for runtime efficiency, leaving room for faster inference and deployment. Second, the real-world evaluation remains limited in scale, requiring larger studies to further validate robustness. Third, extending beyond navigation to domains such as robotic manipulation remains future work.

\section*{Acknowledgements}
This work was supported by the Center of Excellence on New Mobility and Automated Vehicles (Mobility COE), the National Science Foundation (NSF) under Award No. 2346267, POSE: Phase II - DriveX: An Open Source Ecosystem for Automated Driving and Intelligent Transportation Research, and the NVIDIA Academic Grant Program. Yun Zhang was supported by the Amazon Trainium Fellowship.

\section*{Impact Statement}
This work advances VLA models for real-time robot navigation by addressing inference latency and limited onboard computation. TIC-VLA may support deployment in service robotics, logistics, and assisted mobility, especially on edge platforms. At the same time, robots operating in shared human spaces can pose safety and ethical risks, including misinterpretation of instructions, collisions, and socially inappropriate behavior. Although latency-aware reactive control helps reduce these risks, the system remains subject to perception, reasoning, and control failures. Human oversight is therefore necessary in safety-critical settings.


\bibliography{ref}

@article{hirose2025omnivla,
  title={OmniVLA: An Omni-Modal Vision-Language-Action Model for Robot Navigation},
  author={Hirose, Noriaki and Glossop, Catherine and Shah, Dhruv and Levine, Sergey},
  journal={arXiv preprint arXiv:2509.19480},
  year={2025}
}

@inproceedings{bar2025navigation,
  title={Navigation world models},
  author={Bar, Amir and Zhou, Gaoyue and Tran, Danny and Darrell, Trevor and LeCun, Yann},
  booktitle={Proceedings of the Computer Vision and Pattern Recognition Conference},
  pages={15791--15801},
  year={2025}
}

@inproceedings{yuan2025opennav,
  title={Opennav: Open-world navigation with multimodal large language models},
  author={Yuan, Mingfeng and Wang, Letian and Waslander, Steven L},
  booktitle={2025 IEEE/RSJ International Conference on Intelligent Robots and Systems (IROS)},
  pages={18948--18955},
  year={2025},
  organization={IEEE}
}

@article{raychaudhuri2025semantic,
  title={Semantic Mapping in Indoor Embodied AI-A Survey on Advances, Challenges, and Future Directions},
  author={Raychaudhuri, Sonia and Chang, Angel X},
  journal={Transactions on Machine Learning Research},
  year={2025}
}

@inproceedings{liu2025x,
  title={X-mobility: End-to-end generalizable navigation via world modeling},
  author={Liu, Wei and Zhao, Huihua and Li, Chenran and Biswas, Joydeep and Okal, Billy and Goyal, Pulkit and Chang, Yan and Pouya, Soha},
  booktitle={2025 IEEE International Conference on Robotics and Automation (ICRA)},
  pages={7569--7576},
  year={2025},
  organization={IEEE}
}

@article{zhang2024navid,
  title={Navid: Video-based vlm plans the next step for vision-and-language navigation},
  author={Zhang, Jiazhao and Wang, Kunyu and Xu, Rongtao and Zhou, Gengze and Hong, Yicong and Fang, Xiaomeng and Wu, Qi and Zhang, Zhizheng and Wang, He},
  journal={arXiv preprint arXiv:2402.15852},
  year={2024}
}

@article{eftekhar2024one,
  title={The one ring: a robotic indoor navigation generalist},
  author={Eftekhar, Ainaz and Hendrix, Rose and Weihs, Luca and Duan, Jiafei and Caglar, Ege and Salvador, Jordi and Herrasti, Alvaro and Han, Winson and VanderBil, Eli and Kembhavi, Aniruddha and others},
  journal={arXiv preprint arXiv:2412.14401},
  year={2024}
}

@article{dong2025ha,
  title={HA-VLN: A Benchmark for Human-Aware Navigation in Discrete-Continuous Environments with Dynamic Multi-Human Interactions, Real-World Validation, and an Open Leaderboard},
  author={Dong, Yifei and Wu, Fengyi and He, Qi and Li, Heng and Li, Minghan and Cheng, Zebang and Zhou, Yuxuan and Sun, Jingdong and Dai, Qi and Cheng, Zhi-Qi and others},
  journal={arXiv preprint arXiv:2503.14229},
  year={2025}
}

@article{zhang2025flexvln,
  title={Flexvln: Flexible adaptation for diverse vision-and-language navigation tasks},
  author={Zhang, Siqi and Qiao, Yanyuan and Wang, Qunbo and Guo, Longteng and Wei, Zhihua and Liu, Jing},
  journal={IEEE Transactions on Multimedia},
  volume={27},
  pages={6307--6318},
  year={2025},
  publisher={IEEE}
}

@article{zeng2025janusvln,
  title={Janusvln: Decoupling semantics and spatiality with dual implicit memory for vision-language navigation},
  author={Zeng, Shuang and Qi, Dekang and Chang, Xinyuan and Xiong, Feng and Xie, Shichao and Wu, Xiaolong and Liang, Shiyi and Xu, Mu and Wei, Xing},
  journal={arXiv preprint arXiv:2509.22548},
  year={2025}
}

@article{tang2025deep,
  title={Deep reinforcement learning for robotics: A survey of real-world successes},
  author={Tang, Chen and Abbatematteo, Ben and Hu, Jiaheng and Chandra, Rohan and Mart{\'\i}n-Mart{\'\i}n, Roberto and Stone, Peter},
  journal={Annual Review of Control, Robotics, and Autonomous Systems},
  volume={8},
  number={1},
  pages={153--188},
  year={2025},
  publisher={Annual Reviews}
}

@article{du2025vl,
  title={Vl-nav: real-time vision-language navigation with spatial reasoning},
  author={Du, Yi and Fu, Taimeng and Chen, Zhuoqun and Li, Bowen and Su, Shaoshu and Zhao, Zhipeng and Wang, Chen},
  journal={arXiv preprint arXiv:2502.00931},
  year={2025}
}

@misc{InternVLA-N1_2025,
  title        = {InternVLA-N1: An Open Dual-System Navigation Foundation Model with Learned Latent Plans},
  author       = {InternRobotics},
  year         = {2025},
  howpublished = {\url{https://huggingface.co/InternRobotics/InternVLA-N1}},
  note         = {Accessed: 2025-10-10}
}

@inproceedings{zhou2024navgpt,
  title={Navgpt: Explicit reasoning in vision-and-language navigation with large language models},
  author={Zhou, Gengze and Hong, Yicong and Wu, Qi},
  booktitle={Proceedings of the AAAI Conference on Artificial Intelligence},
  volume={38},
  number={7},
  pages={7641--7649},
  year={2024}
}

@article{li2025simplevla,
  title={SimpleVLA-RL: Scaling VLA Training via Reinforcement Learning},
  author={Li, Haozhan and Zuo, Yuxin and Yu, Jiale and Zhang, Yuhao and Yang, Zhaohui and Zhang, Kaiyan and Zhu, Xuekai and Zhang, Yuchen and Chen, Tianxing and Cui, Ganqu and others},
  journal={arXiv preprint arXiv:2509.09674},
  year={2025}
}

@article{schulman2017proximal,
  title={Proximal policy optimization algorithms},
  author={Schulman, John and Wolski, Filip and Dhariwal, Prafulla and Radford, Alec and Klimov, Oleg},
  journal={arXiv preprint arXiv:1707.06347},
  year={2017}
}

@article{lu2025vla,
  title={Vla-rl: Towards masterful and general robotic manipulation with scalable reinforcement learning},
  author={Lu, Guanxing and Guo, Wenkai and Zhang, Chubin and Zhou, Yuheng and Jiang, Haonan and Gao, Zifeng and Tang, Yansong and Wang, Ziwei},
  journal={arXiv preprint arXiv:2505.18719},
  year={2025}
}

@article{wu2023human,
  title={Human-guided reinforcement learning with sim-to-real transfer for autonomous navigation},
  author={Wu, Jingda and Zhou, Yanxin and Yang, Haohan and Huang, Zhiyu and Lv, Chen},
  journal={IEEE Transactions on Pattern Analysis and Machine Intelligence},
  volume={45},
  number={12},
  pages={14745--14759},
  year={2023},
  publisher={IEEE}
}

@article{zhu2025internvl3,
  title={Internvl3: Exploring advanced training and test-time recipes for open-source multimodal models},
  author={Zhu, Jinguo and Wang, Weiyun and Chen, Zhe and Liu, Zhaoyang and Ye, Shenglong and Gu, Lixin and Tian, Hao and Duan, Yuchen and Su, Weijie and Shao, Jie and others},
  journal={arXiv preprint arXiv:2504.10479},
  year={2025}
}

@misc{IsaacSim2025,
  title        = {{IsaacSim: An open-source robotics simulation platform on NVIDIA Omniverse}},
  author       = {{Isaac-sim development team}},
  year         = {2025},
  howpublished = {\url{https://github.com/isaac-sim/IsaacSim}},
  note         = {Accessed: 2025-10-09; version v5.0.0},
}

@inproceedings{yu2025correctnav,
  title={Correctnav: Self-correction flywheel empowers vision-language-action navigation model},
  author={Yu, Zhuoyuan and Long, Yuxing and Yang, Zihan and Zeng, Chengyan and Fan, Hongwei and Zhang, Jiyao and Dong, Hao},
  booktitle={Proceedings of the AAAI Conference on Artificial Intelligence},
  volume={40},
  number={22},
  pages={18737--18745},
  year={2026}
}

@inproceedings{hu2025composablenav,
  title={ComposableNav: Instruction-Following Navigation in Dynamic Environments via Composable Diffusion},
  author={Hu, Zichao and Tang, Chen and Munje, Michael Joseph and Zhu, Yifeng and Liu, Alex and Liu, Shuijing and Warnell, Garrett and Stone, Peter and Biswas, Joydeep},
  booktitle={Conference on Robot Learning},
  pages={4246--4268},
  year={2025},
  organization={PMLR}
}

@article{karnan2022socially,
  title={Socially compliant navigation dataset (scand): A large-scale dataset of demonstrations for social navigation},
  author={Karnan, Haresh and Nair, Anirudh and Xiao, Xuesu and Warnell, Garrett and Pirk, S{\"o}ren and Toshev, Alexander and Hart, Justin and Biswas, Joydeep and Stone, Peter},
  journal={IEEE Robotics and Automation Letters},
  volume={7},
  number={4},
  pages={11807--11814},
  year={2022},
  publisher={IEEE}
}

@inproceedings{liang2025gnd,
  title={Gnd: Global navigation dataset with multi-modal perception and multi-category traversability in outdoor campus environments},
  author={Liang, Jing and Das, Dibyendu and Song, Daeun and Shuvo, Md Nahid Hasan and Durrani, Mohammad and Taranath, Karthik and Penskiy, Ivan and Manocha, Dinesh and Xiao, Xuesu},
  booktitle={2025 IEEE International Conference on Robotics and Automation (ICRA)},
  pages={2383--2390},
  year={2025},
  organization={IEEE}
}

@inproceedings{munje2025socialnav,
  title={SocialNav-SUB: Benchmarking VLMs for Scene Understanding in Social Robot Navigation},
  author={Munje, Michael Joseph and Tang, Chen and Liu, Shuijing and Hu, Zichao and Zhu, Yifeng and Cui, Jiaxun and Warnell, Garrett and Biswas, Joydeep and Stone, Peter},
  booktitle={Conference on Robot Learning},
  pages={1120--1143},
  year={2025},
  organization={PMLR}
}

@article{gao2025octonav,
  title={OctoNav: Towards Generalist Embodied Navigation},
  author={Gao, Chen and Jin, Liankai and Peng, Xingyu and Zhang, Jiazhao and Deng, Yue and Li, Annan and Wang, He and Liu, Si},
  journal={arXiv preprint arXiv:2506.09839},
  year={2025}
}

@inproceedings{zhao2025cot,
  title={Cot-vla: Visual chain-of-thought reasoning for vision-language-action models},
  author={Zhao, Qingqing and Lu, Yao and Kim, Moo Jin and Fu, Zipeng and Zhang, Zhuoyang and Wu, Yecheng and Li, Zhaoshuo and Ma, Qianli and Han, Song and Finn, Chelsea and others},
  booktitle={Proceedings of the Computer Vision and Pattern Recognition Conference},
  pages={1702--1713},
  year={2025}
}

@article{lin2025onetwovla,
  title={OneTwoVLA: A Unified Vision-Language-Action Model with Adaptive Reasoning},
  author={Lin, Fanqi and Nai, Ruiqian and Hu, Yingdong and You, Jiacheng and Zhao, Junming and Gao, Yang},
  journal={arXiv preprint arXiv:2505.11917},
  year={2025}
}

@inproceedings{xu2024mobility,
  title={Mobility VLA: Multimodal instruction navigation with long-context VLMs and topological graphs},
  author={Xu, Zhuo and Chiang, Hao-Tien Lewis and Fu, Zipeng and Jacob, Mithun George and Zhang, Tingnan and Lee, Tsang-Wei Edward and Yu, Wenhao and Schenck, Connor and Rendleman, David and Shah, Dhruv and others},
  booktitle={8th Annual Conference on Robot Learning},
  year={2024}
}

@article{elnoor2025vi,
  title={ViLAM: Distilling Vision-Language Reasoning into Attention Maps for Social Robot Navigation}, 
  author={Elnoor, Mohamed and Weerakoon, Kasun and Seneviratne, Gershom and Liang, Jing and Rajagopal, Vignesh and Manocha, Dinesh},
  journal={arXiv preprint arXiv:2503.09820},
  year={2026}
}

@article{cai2025navdp,
  title={NavDP: Learning Sim-to-Real Navigation Diffusion Policy with Privileged Information Guidance},
  author={Cai, Wenzhe and Peng, Jiaqi and Yang, Yuqiang and Zhang, Yujian and Wei, Meng and Wang, Hanqing and Chen, Yilun and Wang, Tai and Pang, Jiangmiao},
  journal={arXiv preprint arXiv:2505.08712},
  year={2025}
}

@inproceedings{krantz_vlnce_2020,
  title={Beyond the Nav-Graph: Vision and Language Navigation in Continuous Environments},
  author={Jacob Krantz and Erik Wijmans and Arjun Majundar and Dhruv Batra and Stefan Lee},
  booktitle={European Conference on Computer Vision (ECCV)},
  year={2020}
 }

@article{he2025seeing,
  title={From seeing to experiencing: Scaling navigation foundation models with reinforcement learning},
  author={He, Honglin and Ma, Yukai and Wu, Wayne and Zhou, Bolei},
  journal={arXiv preprint arXiv:2507.22028},
  year={2025}
}

@inproceedings{ku2020room,
  title={Room-Across-Room: Multilingual Vision-and-Language Navigation with Dense Spatiotemporal Grounding},
  author={Ku, Alexander and Anderson, Peter and Patel, Roma and Ie, Eugene and Baldridge, Jason},
  booktitle={Proceedings of the 2020 Conference on Empirical Methods in Natural Language Processing (EMNLP)},
  pages={4392--4412},
  year={2020}
}

@article{zhang2025embodied,
  title={Embodied navigation foundation model},
  author={Zhang, Jiazhao and Li, Anqi and Qi, Yunpeng and Li, Minghan and Liu, Jiahang and Wang, Shaoan and Liu, Haoran and Zhou, Gengze and Wu, Yuze and Li, Xingxing and others},
  journal={arXiv preprint arXiv:2509.12129},
  year={2025}
}

@article{liu2025compass,
  title={COMPASS: Cross-embodiment Mobility Policy via Residual RL and Skill Synthesis},
  author={Liu, Wei and Zhao, Huihua and Li, Chenran and Biswas, Joydeep and Pouya, Soha and Chang, Yan},
  journal={arXiv preprint arXiv:2502.16372},
  year={2025}
}

@article{shukor2025smolvla,
  title={Smolvla: A vision-language-action model for affordable and efficient robotics},
  author={Shukor, Mustafa and Aubakirova, Dana and Capuano, Francesco and Kooijmans, Pepijn and Palma, Steven and Zouitine, Adil and Aractingi, Michel and Pascal, Caroline and Russi, Martino and Marafioti, Andres and others},
  journal={arXiv preprint arXiv:2506.01844},
  year={2025}
}

@inproceedings{payandeh2024social,
  title={Social-LLaVA: Enhancing Social Robot Navigation through Human-Language Reasoning},
  author={Payandeh, Amirreza and Song, Daeun and Nazeri, Mohammad and Liang, Jing and Mukherjee, Praneel and Raj, Amir Hossain and Kong, Yangzhe and Manocha, Dinesh and Xiao, Xuesu},
  booktitle={2025 IEEE/RSJ International Conference on Intelligent Robots and Systems (IROS)},
  pages={17192--17198},
  year={2025},
  organization={IEEE}
}

@inproceedings{shcherbyna2025arena,
  title={Arena 4.0: A comprehensive ros2 development and benchmarking platform for human-centric navigation using generative-model-based environment generation},
  author={Shcherbyna, Volodymyr and Kastner, Linh and Diaz, Diego and Nguyen, Huu Giang and Schreff, Maximilian Ho--Kyoung and Seeger, Tim and Kreutz, Jonas and Martban, Ahmed and Shen, Zhengcheng and Zeng, Huajian and others},
  booktitle={2025 IEEE International Conference on Robotics and Automation (ICRA)},
  pages={9138--9144},
  year={2025},
  organization={IEEE}
}

@article{wei2025streamvln,
  title={Streamvln: Streaming vision-and-language navigation via slowfast context modeling},
  author={Wei, Meng and Wan, Chenyang and Yu, Xiqian and Wang, Tai and Yang, Yuqiang and Mao, Xiaohan and Zhu, Chenming and Cai, Wenzhe and Wang, Hanqing and Chen, Yilun and others},
  journal={arXiv preprint arXiv:2507.05240},
  year={2025}
}

@inproceedings{wu2025towards,
  title={Towards autonomous micromobility through scalable urban simulation},
  author={Wu, Wayne and He, Honglin and Zhang, Chaoyuan and He, Jack and Zhao, Seth Z and Gong, Ran and Li, Quanyi and Zhou, Bolei},
  booktitle={Proceedings of the Computer Vision and Pattern Recognition Conference},
  pages={27553--27563},
  year={2025}
}

@inproceedings{wang2025trackvla,
  title={TrackVLA: Embodied Visual Tracking in the Wild},
  author={Wang, Shaoan and Zhang, Jiazhao and Li, Minghan and Liu, Jiahang and Li, Anqi and Wu, Kui and Zhong, Fangwei and Yu, Junzhi and Zhang, Zhizheng and Wang, He},
  booktitle={Conference on Robot Learning},
  pages={4139--4164},
  year={2025},
  organization={PMLR}
}

@inproceedings{choi2024embodied,
  title={Embodied CoT distillation from LLM to off-the-shelf agents},
  author={Choi, Wonje and Kim, Woo Kyung and Yoo, Minjong and Woo, Honguk},
  booktitle={Proceedings of the 41st International Conference on Machine Learning},
  pages={8702--8721},
  year={2024}
}

@article{zhang2024uni,
  title={Uni-navid: A video-based vision-language-action model for unifying embodied navigation tasks},
  author={Zhang, Jiazhao and Wang, Kunyu and Wang, Shaoan and Li, Minghan and Liu, Haoran and Wei, Songlin and Wang, Zhongyuan and Zhang, Zhizheng and Wang, He},
  journal={arXiv preprint arXiv:2412.06224},
  year={2024}
}

@article{hu2025astranav,
  title={AstraNav-World: World Model for Foresight Control and Consistency},
  author={Hu, Junjun and Chen, Jintao and Bai, Haochen and Luo, Minghua and Xie, Shichao and Chen, Ziyi and Liu, Fei and Chu, Zedong and Xue, Xinda and Ren, Botao and others},
  journal={arXiv preprint arXiv:2512.21714},
  year={2025}
}

@article{bai2025qwen2,
  title={Qwen2.5-VL technical report},
  author={Bai, Shuai and Chen, Keqin and Liu, Xuejing and Wang, Jialin and Ge, Wenbin and Song, Sibo and Dang, Kai and Wang, Peng and Wang, Shijie and Tang, Jun and others},
  journal={arXiv preprint arXiv:2502.13923},
  year={2025}
}

@article{mittal2025isaaclab,
  title={Isaac Lab: A GPU-accelerated simulation framework for multi-modal robot learning},
  author={Mittal, Mayank and Roth, Pascal and Tigue, James and Richard, Antoine and Zhang, Octi and Du, Peter and Serrano-Mu{\~n}oz, Antonio and Yao, Xinjie and Zurbr{\"u}gg, Ren{\'e} and Rudin, Nikita and others},
  journal={arXiv preprint arXiv:2511.04831},
  year={2025}
}

@article{wei2025ground,
  title={Ground Slow, Move Fast: A Dual-System Foundation Model for Generalizable Vision-and-Language Navigation},
  author={Wei, Meng and Wan, Chenyang and Peng, Jiaqi and Yu, Xiqian and Yang, Yuqiang and Feng, Delin and Cai, Wenzhe and Zhu, Chenming and Wang, Tai and Pang, Jiangmiao and others},
  journal={arXiv preprint arXiv:2512.08186},
  year={2025}
}

@article{huang2025mobilevla,
  title={MobileVLA-R1: Reinforcing Vision-Language-Action for Mobile Robots},
  author={Huang, Ting and Li, Dongjian and Yang, Rui and Zhang, Zeyu and Yang, Zida and Tang, Hao},
  journal={arXiv preprint arXiv:2511.17889},
  year={2025}
}

@article{chen2025socialnav,
  title={SocialNav: Training Human-Inspired Foundation Model for Socially-Aware Embodied Navigation},
  author={Chen, Ziyi and Guo, Yingnan and Chu, Zedong and Luo, Minghua and Shen, Yanfen and Sun, Mingchao and Hu, Junjun and Xie, Shichao and Yang, Kuan and Shi, Pei and others},
  journal={arXiv preprint arXiv:2511.21135},
  year={2025}
}

@inproceedings{chandaka2025human,
  title={Human-like Navigation in a World Built for Humans},
  author={Chandaka, Bhargav and Wang, Gloria Xinyue and Chen, Haozhe and Che, Henry and Zhai, Albert J and Wang, Shenlong},
  booktitle={Conference on Robot Learning},
  pages={1790--1808},
  year={2025},
  organization={PMLR}
}

@article{castro2025vamos,
  title={VAMOS: A Hierarchical Vision-Language-Action Model for Capability-Modulated and Steerable Navigation},
  author={Castro, Mateo Guaman and Rajagopal, Sidharth and Gorbatov, Daniel and Schmittle, Matt and Baijal, Rohan and Zhang, Octi and Scalise, Rosario and Talia, Sidharth and Romig, Emma and de Melo, Celso and others},
  journal={arXiv preprint arXiv:2510.20818},
  year={2025}
}

@article{driess2025knowledge,
  title={Knowledge insulating vision-language-action models: Train fast, run fast, generalize better},
  author={Driess, Danny and Springenberg, Jost and Ichter, Brian and Yu, Lili and Li-Bell, Adrian and Pertsch, Karl and Ren, Allen and Walke, Homer and Vuong, Quan and Shi, Lucy Xiaoyang and others},
  journal={Advances in Neural Information Processing Systems},
  volume={38},
  pages={102867--102888},
  year={2026}
}

@article{chen2025era,
  title={Era: Transforming VLMs into embodied agents via embodied prior learning and online reinforcement learning},
  author={Chen, Hanyang and Zhao, Mark and Yang, Rui and Ma, Qinwei and Yang, Ke and Yao, Jiarui and Wang, Kangrui and Bai, Hao and Wang, Zhenhailong and Pan, Rui and others},
  journal={arXiv preprint arXiv:2510.12693},
  year={2025}
}

@article{cheng2024navila,
  title={Navila: Legged robot vision-language-action model for navigation},
  author={Cheng, An-Chieh and Ji, Yandong and Yang, Zhaojing and Gongye, Zaitian and Zou, Xueyan and Kautz, Jan and B{\i}y{\i}k, Erdem and Yin, Hongxu and Liu, Sifei and Wang, Xiaolong},
  journal={arXiv preprint arXiv:2412.04453},
  year={2024}
}

@article{zhou2025autovla,
  title={Autovla: A vision-language-action model for end-to-end autonomous driving with adaptive reasoning and reinforcement fine-tuning},
  author={Zhou, Zewei and Cai, Tianhui and Zhao, Seth and Zhang, Yun and Huang, Zhiyu and Zhou, Bolei and Ma, Jiaqi},
  journal={Advances in Neural Information Processing Systems},
  volume={38},
  pages={27920--27956},
  year={2026}
}

@article{payandeh2025narrate2nav,
  title={Narrate2Nav: Real-Time Visual Navigation with Implicit Language Reasoning in Human-Centric Environments},
  author={Payandeh, Amirreza and Pokhrel, Anuj and Song, Daeun and Zampieri, Marcos and Xiao, Xuesu},
  journal={arXiv preprint arXiv:2506.14233},
  year={2025}
}

@inproceedings{li2019robust,
  title={Robust navigation with language pretraining and stochastic sampling},
  author={Li, Xiujun and Li, Chunyuan and Xia, Qiaolin and Bisk, Yonatan and Celikyilmaz, Asli and Gao, Jianfeng and Smith, Noah A and Choi, Yejin},
  booktitle={Proceedings of the 2019 Conference on Empirical Methods in Natural Language Processing and the 9th International Joint Conference on Natural Language Processing (EMNLP-IJCNLP)},
  pages={1494--1499},
  year={2019}
}

@inproceedings{gong2024cognition,
  title={From cognition to precognition: A future-aware framework for social navigation},
  author={Gong, Zeying and Hu, Tianshuai and Qiu, Ronghe and Liang, Junwei},
  booktitle={2025 IEEE International Conference on Robotics and Automation (ICRA)},
  pages={9122--9129},
  year={2025},
  organization={IEEE}
}

@article{grutopia,
  title={Grutopia: Dream general robots in a city at scale},
  author={Wang, Hanqing and Chen, Jiahe and Huang, Wensi and Ben, Qingwei and Wang, Tai and Mi, Boyu and Huang, Tao and Zhao, Siheng and Chen, Yilun and Yang, Sizhe and others},
  journal={arXiv preprint arXiv:2407.10943},
  year={2024}
}

@inproceedings{wang2025rethinking,
  title={Rethinking the embodied gap in vision-and-language navigation: A holistic study of physical and visual disparities},
  author={Wang, Liuyi and Xia, Xinyuan and Zhao, Hui and Wang, Hanqing and Wang, Tai and Chen, Yilun and Liu, Chengju and Chen, Qijun and Pang, Jiangmiao},
  booktitle={Proceedings of the IEEE/CVF International Conference on Computer Vision},
  pages={9455--9465},
  year={2025}
}
\bibliographystyle{icml2026}

\newpage
\appendix
\onecolumn
\begin{center}
{\LARGE \textbf{\textit{TIC-VLA} Supplementary Materials}}   
\vspace{0.4cm}
\end{center}

\section{Model Details}

\textbf{VLM Backbone.}
Our model is built upon an \textit{InternVL3-1B} vision-language model, which consists of an \textit{InternViT-300M} vision encoder and a \textit{Qwen2.5-0.5B} language model. The VLM processes a sequence of historical images spanning up to a nine-second temporal window to perform reasoning and generate high-level planning waypoints. Images are sampled at three-second intervals, resulting in three historical frames and one current frame. Each image is captured at a resolution of \(1920\times1080\), and these images are encoded into visual tokens of size $4 \times 256$, which together provide both temporal and historical context for reasoning.

Due to reasoning inference latency, the visual observations fed into the VLM are temporally lagged relative to the real-time sensory inputs received by the action expert. To ensure consistent visual representations across modules, the vision encoder is shared between the VLM and the action expert. Specifically, the shared encoder processes the current image observation and outputs visual tokens that are directly consumed by the action expert for real-time control.

\textbf{Action Expert.}
The action expert is implemented as a Transformer-based architecture composed of six cross-attention layers. Its inputs include: (1) current visual tokens (256 tokens) extracted from the shared vision encoder, (2) the robot’s proprioceptive state, (3) delayed semantic guidance from the VLM,  and (4) latency-related metadata. The VLM guidance is provided either in the form of abstract waypoint predictions or as key-value (KV) cache features from the final Transformer layer of the VLM. The robot state is represented by a vector consisting of the linear velocities $(v_x, v_y)$ and the yaw rate ($\omega_z$). 

To explicitly account for inference latency, we incorporate latency-related metadata, including the estimated robot displacement during the inference delay $(\Delta x, \Delta y, \Delta \theta)$, and the corresponding time delay $\Delta t$. Visual tokens are projected from 896 dimensions to 512 dimensions, VLM KV cache features are projected from 256 to 512 dimensions, and low-dimensional inputs (waypoint guidance and latency metadata) are projected to 512 dimensions, followed by positional embeddings. We apply token dropout to the KV-cache tokens (drop rate 0.1) to encourage robustness to missing semantic context, and apply dropout to the encoded robot-state tokens/positional embeddings (drop rate 0.5) to reduce overfitting to state inputs.

All processed features are treated as scene context and serve as keys and values in the cross-attention Transformer. The queries are learnable embeddings indexed by time steps, enabling the model to produce structured predictions. Transformer outputs are passed through an MLP to produce action chunks at each predicted step.

We predict actions over a future horizon of three seconds, discretized into $T=30$ action chunks. Each action is parameterized as $(d_x, d_y, d_\theta)$, representing relative motion increments. A simple integrator (cumulative sum) is applied to accumulate these actions into a continuous trajectory, yielding waypoints of the form $(x, y, \theta)$.

\textbf{Value Network.}
The value network shares the same vision encoder as the action expert and operates on the current image observation. The resulting 256 visual tokens are first processed by three 1D convolutional layers followed by an average pooling layer, reducing them to a compact feature representation. In parallel, the robot state $(v_x, v_y)$ and yaw rate ($\omega_z$), as well as the relative goal position, are concatenated and projected into the same feature dimension using MLPs. These feature representations are concatenated and passed through a final MLP to predict the scalar value estimate. 

The hyperparameters used in model components are summarized in \cref{tab:arch}.

\begin{table}[ht]
    \centering
    \begin{minipage}{0.32\columnwidth}
        \centering
        \caption{Model Parameters}
        \begin{tabular}{l c}
            \toprule
            \textbf{Parameter} & \textbf{Value} \\
            \midrule
            Token dim. & 896 \\
            Proj. dim. & 512 \\
            Dropout rate-KV cache & 0.1 \\
            Dropout rate-state & 0.5 \\
            Cross-attn layers & 6 \\
            Query dim. & 512 \\
            Queries ($T$) & 30 \\
            FFN dim. & 2048 \\
            Horizon (s) & 3 \\
            Conv1D layers & 3 \\
            \bottomrule
        \end{tabular}
        \label{tab:arch}
    \end{minipage}\hfill
    \begin{minipage}{0.32\columnwidth}
        \centering
        \caption{Training Parameters}
        \begin{tabular}{l c}
            \toprule
            \textbf{Parameter} & \textbf{Value} \\
            \midrule
            LR-Action & $2\!\times\!10^{-4}$ \\
            LR-VLM  & $2\!\times\!10^{-5}$ \\
            Weight decay & $1\!\times\!10^{-2}$ \\
            Batch size-VLM & 2 per GPU  \\
            Batch size-Action & 16 per GPU \\
            Epoch & 10 \\
            LR sched. & Cosine \\
            Warmup & 1000 step \\
            Grad. clip & 1.0 \\
            \bottomrule
        \end{tabular}
        \label{tab:train}
    \end{minipage}\hfill
    \begin{minipage}{0.32\columnwidth}
        \centering
        \caption{RL Parameters  }
        \begin{tabular}{l c}
            \toprule
            \textbf{Parameter} & \textbf{Value} \\
            \midrule
            Action LR & $1\!\times\!10^{-5}$ \\
            Value LR & $1\!\times\!10^{-4}$ \\
            Rollout steps & 1024 \\
            Gamma & 0.99 \\
            GAE Lambda & 0.95 \\
            Clip ratio & 0.2 \\
            Grad. clip & 0.5 \\
            Entropy coef. & 0.001 \\
            PPO epochs & 5 \\
            Batch size & 64 \\
            \bottomrule
        \end{tabular}
        \label{tab:rl}
    \end{minipage}
\end{table}

\section{Training Details}

\textbf{Data Processing.}
We start from long-horizon raw data collected continuously within a single scenario or location. Each raw sequence is segmented into a set of fixed-length episodes, each spanning 20 seconds. For each episode, we retain only the egocentric RGB images and the corresponding robot pose trajectories, discarding other sensor modalities. All data are temporally downsampled to 10~Hz to ensure a consistent representation across datasets and training stages.

For each episode, we construct a structured representation consisting of a history window, a current frame, and a future trajectory. The history contains past frames with relative pose offsets, while the future trajectory is represented as a sequence of relative motion offsets with respect to the current pose. These offsets are expressed in the robot-centric coordinate frame. Episodes are further subsampled at a fixed temporal stride to reduce redundancy, resulting in one annotated sample approximately every three seconds. This preprocessing yields a compact yet diverse set of training samples suitable for both vision-language modeling and control learning.

\begin{figure}[ht]
    \centering
    \includegraphics[width=\linewidth]{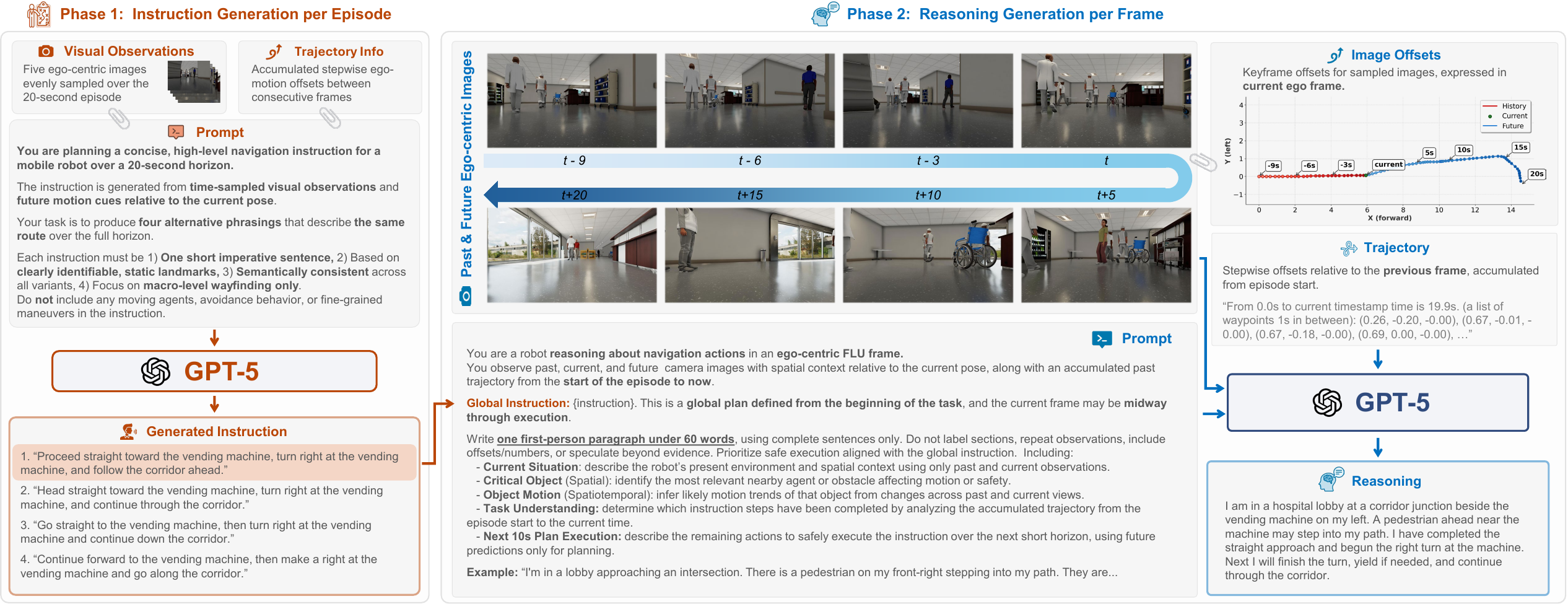}
    \caption{Overview of the annotation pipeline for VLM SFT. Representative frames and future trajectory information are used to generate long-horizon navigation instructions and concise CoT reasoning annotations.}
    \label{fig:annotate}
\end{figure}

\begin{figure}[ht]
    \centering
    \includegraphics[width=\linewidth]{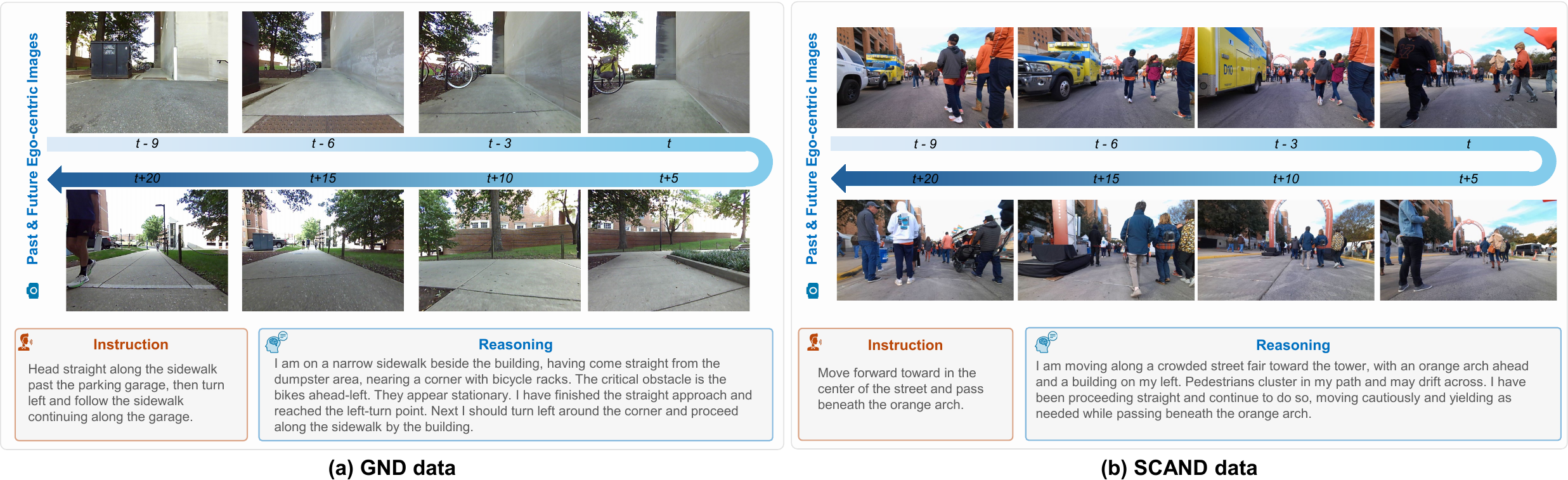}
    \caption{Examples of navigation instruction and CoT reasoning annotations from the GND and SCAND datasets.}
    \label{fig:example}
\end{figure}

\textbf{VLM Supervised Fine-Tuning.}
We first perform supervised fine-tuning (SFT) of the VLM. For each 20-second episode, we automatically generate a long-term navigation instruction using GPT-5. To provide sufficient temporal context, we select five representative frames corresponding to the current time step and future time offsets at 5, 10, 15, and 20 seconds. These frames, together with the future trajectory offsets, are used to prompt the language model to generate four semantically equivalent instruction variants describing the same route. The variants differ only in wording and are randomly assigned across episodes to increase linguistic diversity while preserving route consistency. \cref{fig:annotate} illustrates the annotation pipeline for navigation instruction generation and scenario reasoning on the DynaNav simulation data, while \cref{fig:example} presents representative examples of instruction and reasoning annotations from the GND and SCAND datasets.

In addition to language instructions, we annotate each episode with a concise reasoning trace. The reasoning trace is generated by conditioning GPT-5 on a richer temporal context that includes the past trajectory, observations at 3, 6, and 9 seconds in the past, the current observation, and future observations.  The past trajectory is summarized at 1s intervals from the start of the episode to the current time, expressed in the local frame of the previous 1s observation, to help the model infer motion patterns and task progress. The generated CoT focuses on five aspects: (1) the current situation, (2) critical objects, (3) object motion, (4) task progress relative to the instruction, and (5) the next short-horizon plan. To ensure efficiency and consistency, the reasoning trace is constrained to be brief and is stored alongside the episode.

During SFT, the vision encoder is frozen. The language model is trained to autoregressively generate either (1) a reasoning-augmented sequence followed by waypoint predictions or (2) waypoint-only outputs. Mixing these two target formats allows the VLM to flexibly produce explicit reasoning when required and output only waypoints to maintain efficiency.

\textbf{Imitation Learning with Delayed Inference.}
After fine-tuning the VLM, we train the action expert via imitation learning while explicitly accounting for inference latency. To simulate real-world deployment conditions, we randomly sample inference delays and expose the action policy to temporally delayed VLM inference hidden states together with explicit latency metadata. During training, the policy is conditioned on delayed ground-truth reasoning annotations and guidance waypoints, ensuring access to the correct high-level intent. The overall training procedure is summarized in \cref{alg:il}.

\begin{algorithm}
\caption{Imitation Learning with Delayed Inference}
\label{alg:il}
\begin{algorithmic}[1]
\REQUIRE Demonstration dataset $\mathcal{D}$, action policy $\pi_\theta$, frozen VLM $f_{\text{LM}}$
\FOR{each training iteration}
    \STATE Sample instruction, current visual observation, and expert future trajectory, $(I, x_t, \{\hat{\mathbf{p}}_t\}) \sim \mathcal{D}$
    \STATE Sample inference delay $\Delta t \sim U(0,10)$
    \STATE Retrieve delayed visual observations and reasoning traces $(\mathcal{X}^{\text{vlm}}_{t-\Delta t}, \mathcal{R}_{t-\Delta t}) \sim \mathcal{D}$
    \STATE Obtain delayed VLM hidden state (KV cache) $S_{t-\Delta t} \leftarrow f_{\text{LM}}(\mathcal{V}^{\text{vlm}}_{t-\Delta t}, I, \mathcal{R}_{t-\Delta t})$
    \STATE Extract current visual tokens $\mathcal{V}^{\text{act}}_t=\text{Enc}_{\text{vis}}(x_{t})$
    \STATE Encode robot state $s_t$ and latency metadata $(\Delta t, \Delta x, \Delta y, \Delta \theta)$
    \STATE Predict action chunk $\{a_t^{(1:T)}\} \leftarrow \pi_\theta(\mathcal{V}^{\text{act}}_t, \mathcal{S}_{t-\Delta t}, s_t, \Delta t, \Delta x, \Delta y, \Delta \theta)$
    \STATE Integrate the predicted actions to obtain trajectory $\{\mathbf{p}_t^{(1:T)}\}$
    \STATE Compute imitation loss $\mathcal{L}_{a}$
    \STATE Update $\theta$ by minimizing $\mathcal{L}_{a}$
\ENDFOR
\end{algorithmic}
\end{algorithm}

For the waypoint-guided policy variant, we transform the delayed VLM-predicted waypoints into the coordinate frame of the robot’s current position before conditioning the action policy. The same transformation is applied consistently during both training and inference. The parameters used for training are provided in \cref{tab:train}.

\textbf{Online Reinforcement Learning.} 
At each control step, the action expert predicts a short-horizon trajectory, from which we select a target corresponding to one second into the future. This target is converted into a desired linear velocity and angular velocity, which define the mean of a Gaussian action distribution. The policy additionally maintains a learnable log-standard-deviation parameter, yielding a stochastic Gaussian policy suitable for on-policy optimization.
To preserve consistency with deployment conditions, the policy continues to operate on temporally delayed VLM inference hidden states and latency metadata during reinforcement learning. Delayed VLM hidden states are cached during rollout collection and reused during policy optimization. We adopt Proximal Policy Optimization (PPO) \cite{schulman2017proximal} with a learned value function to optimize the policy under this delayed-inference setting. The overall procedure is summarized in \cref{alg:rl} and the parameters used for PPO are provided in \cref{tab:rl}.

\begin{algorithm}
\caption{Online Reinforcement Learning}
\label{alg:rl}
\begin{algorithmic}[1]
\REQUIRE Action policy $\pi_\theta$, value network $V_\psi$, frozen VLM $f_{\text{LM}}$, environment $\mathcal{E}$
\FOR{each PPO iteration}
\STATE Initialize rollout buffer $\mathcal{B} \leftarrow \emptyset$
\FOR{each environment step $t$}
\STATE Observe current visual observation $x_t$ and robot state $s_t$
\STATE Run VLM inference asynchronously in the background
\STATE Retrieve delayed VLM inference hidden state $S_{t-\Delta t}$ and latency metadata $(\Delta t, \Delta x, \Delta y, \Delta \theta)$
\STATE Extract current visual tokens $\mathcal{V}^{\text{act}}_t=\text{Enc}_{\text{vis}}(x_{t})$
\STATE Compute Gaussian policy parameters $(\mu_t, \log \sigma) \leftarrow \pi_\theta(\mathcal{V}^{\text{act}}, S_{t-\Delta t}, s_t, \Delta t, \Delta x, \Delta y, \Delta \theta)$
\STATE Sample action $a_t \sim \mathcal{N}(\mu_t, \sigma^2)$
\STATE Execute $a_t$ in environment $\mathcal{E}$ and observe reward $r_t$ and next state
\STATE Store data $(x_t, S_{t-\Delta t}, s_t, a_t, r_t, \Delta t, \Delta x, \Delta y, \Delta \theta)$ in $\mathcal{B}$
\ENDFOR
\STATE Compute returns $\hat{R}_t$ and advantages $\hat{A}_t$ using value network $V_\psi$
\FOR{each PPO optimization epoch}
\STATE Update policy $\pi_\theta$ using clipped surrogate objective
\STATE Update value network $V_\psi$ by minimizing value regression loss
\ENDFOR
\ENDFOR
\end{algorithmic}
\end{algorithm}

\section{Benchmark Details}

\begin{figure}[ht]
    \centering
    \includegraphics[width=\linewidth]{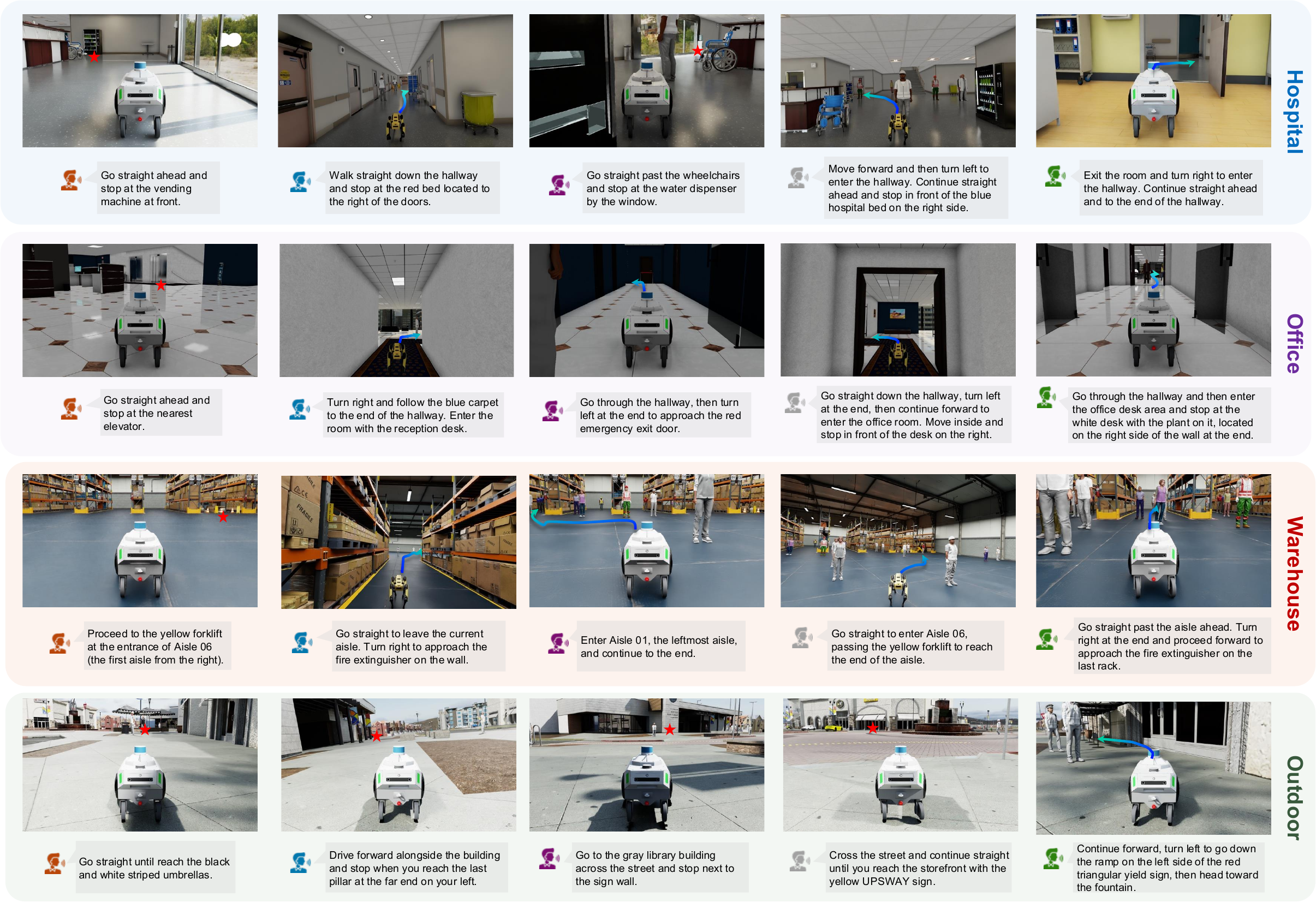}
    \caption{Overview of the DynaNav benchmark. Task instructions are provided, as well as corresponding navigation scenarios across the Hospital, Office, Outdoor, and Warehouse environments, highlighting variations in scene layout, landmarks, and human density.}
    \label{fig:benchmark}
\end{figure}

We evaluate navigation methods using a language-conditioned physics-based benchmark consisting of 85 episodes across four environments: \emph{Hospital}, \emph{Office}, \emph{Outdoor}, and \emph{Warehouse}. Each episode specifies a unique combination of scene layout, start pose, goal location, and natural-language instruction. Task difficulty is systematically increased by varying pedestrian density (from 0 to 200 agents), instruction complexity, and time limits, while maintaining consistent goals within each task. Pedestrian agents exhibit diverse visual appearances and body types to increase visual variability and reduce reliance on appearance-specific cues. \cref{fig:benchmark} illustrates navigation instructions and scenarios included in the benchmark.

For each environment, we design multiple navigation tasks that share the same semantic objective but differ in crowd density, robot platform (Nova Carter or Spot), and timeout constraints. This design enables controlled evaluation of robustness to dynamic obstacles, long-horizon instruction following, and cross-robot generalization. An episode is considered successful if the robot reaches the target location within a predefined distance threshold before the time limit; outdoor environments use a relaxed threshold to account for large-scale spatial uncertainty. All experiments are conducted with a fixed random seed to ensure reproducibility.

\textbf{Hospital Environment.}
We conducted 25 episodes in a large hospital scene featuring long corridors, patient rooms, medical equipment, and service areas. Tasks involve reaching vending machines, hospital beds, water dispensers, and hallway endpoints under dense pedestrian traffic. The hospital setting is characterized by narrow passages, frequent occlusions, and socially constrained motion, making it particularly challenging for safe navigation under crowd interference. For each task type, we systematically vary the number of simulated pedestrians from 0 to 60 and proportionally increase the episode timeout. This setting evaluates robustness to crowd density and social navigation challenges in structured indoor spaces.

\textbf{Office Environment.}
We perform 25 episodes in a complex office environment composed of interconnected rooms, corridors, elevators, and desk areas. Tasks emphasize multi-step spatial instructions involving turns, room entries, and landmark-based goals (e.g., desks, carpets, doors), often beyond immediate perceptual range. The office environment introduces greater topological diversity and branching layouts, requiring precise instruction grounding rather than corridor following. Pedestrian density is again increased from 0 to 60 to study performance degradation under dynamic human participants in semantically rich but spatially compact environments.

\textbf{Warehouse Environment.}
We evaluate 25 episodes in a structured warehouse scene with long aisles, racks, forklifts, cones, and fire extinguishers. The tasks emphasize fine-grained spatial reasoning, such as selecting the correct aisle, identifying object instances in repetitive layouts, and navigating to precise rack locations. The warehouse presents high visual repetition with minimal semantic diversity, stressing the model’s ability to disambiguate similar structures under motion and crowd interference. Pedestrian density is varied from 0 to 60 with adjusted timeouts to control task difficulty. Owing to the warehouse’s compact spatial layout relative to the office and hospital environments, this setting represents the most crowded and visually ambiguous scenario.

\textbf{Outdoor Environment.}
We include 10 episodes in a large-scale outdoor urban environment with uneven terrain, ramps, streets, and large open spaces. These tasks involve long-distance navigation guided by distant landmarks, such as buildings, umbrellas, fountains, and storefronts. The number of pedestrians ranges from 100 to 200, and a relaxed success threshold is used to account for localization uncertainty and large-scale goals. Due to the substantially higher rendering and simulation cost of large outdoor scenes, we limit this setting to 10 episodes while maintaining high environmental diversity. These episodes test scalability, visual clutter robustness, and instruction grounding in outdoor settings, representing the most demanding setting in terms of perception, planning horizon, and real-time control stability.

\section{Experiment Details}

\textbf{Data Collection.}  
We collect demonstration data by asking a human expert to teleoperate the robot using a keyboard across four designed environments (warehouse, office, hospital, and outdoor) within the DynaNav simulation. Each environment contains dynamic human participants. The expert navigates the robot toward predefined goals, and upon completing each demonstration, records a corresponding natural language instruction describing the navigation task. The resulting dataset consists of RGB images and trajectories of robot camera poses. In total, we collected 310 episodes, corresponding to 5.1 hours of navigation data.

\begin{figure}[ht]
    \centering
    \includegraphics[width=0.85\linewidth]{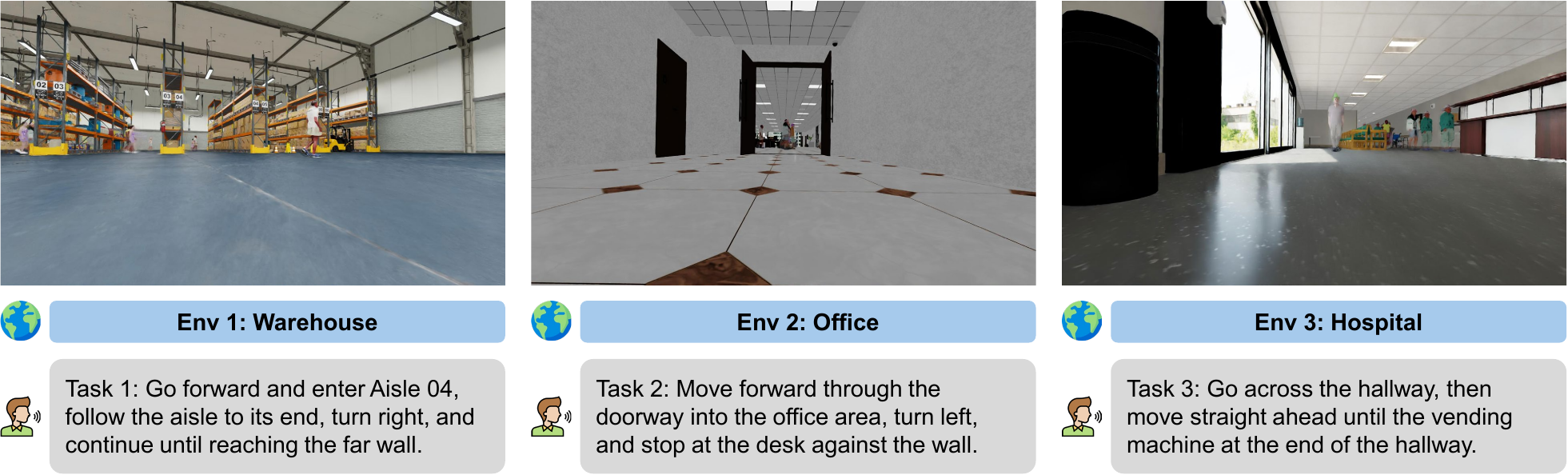}
    \caption{Online reinforcement learning tasks used to train TIC-VLA across three environments and tasks.}
    \label{fig:rl_task}
\end{figure}

\textbf{Online Reinforcement Learning.}  
We perform online RL of the TIC-VLA model on three different tasks (different from testing) across three environments: office, hospital, and warehouse. The environment operates at a control frequency of 10~Hz, corresponding to an environment step of 0.1 seconds. During training, the VLM runs inference asynchronously in the background upon receiving new observations and returns its outputs to the action policy once available. We finetune only the final cross-attention Transformer layer and the MLP decoder of the action expert. To emulate deployment on resource-constrained hardware, we explicitly inject random inference delays before releasing the VLM outputs to the policy, with an average delay of approximately 50 environment steps (5 seconds). The three tasks and environments, illustrated in \cref{fig:rl_task}, are rotated every 100 PPO iterations.

The weights for the reward function (\cref{reward}) are set to $w_g = 400$ for reaching the goal, $w_p = 5$ for progress toward the goal, $w_c = -100$ for collisions with obstacles and humans, and $w_s = -0.1$ for the speed penalty. The target speed for the speed penalty is set to 1~m/s.

\textbf{Baseline Methods.}  
We compare our approach against several baseline methods. The behavior cloning (BC) baseline is trained directly on the demonstration dataset without modeling inference latency or delayed perception, while being provided with privileged access to the relative goal position. For the RL baseline, we train a PPO agent whose policy shares the same architecture as the value network, except that it directly outputs the action mean and standard deviation. The PPO policy is trained separately for each environment using 10 parallel environments and a total of 1M steps per environment. For other baselines, including Uni-NaVid, NAVILA, NavDP, DualVLN, MobileVLA, and OmniVLA, we follow their respective open-source implementations and training.

\textbf{Ablation Action Policy Architectures.}  
We implement the diffusion-based action policy as a conditional denoising model over fixed-length action chunks of $T=30$ steps. The forward diffusion process uses $K=100$ noise steps with a cosine variance schedule, and the denoiser is a Transformer backbone aligned with the query-based policy to ensure comparable capacity. Noisy action tokens are augmented with timestep embeddings and positional encodings, while conditioning information from the vision encoder and cached VLM hidden states is injected at every denoising layer via cross-attention. Actions are normalized prior to training, and the model is trained with the standard denoising mean-squared-error objective to predict clean actions. During inference, we apply DDIM sampling with 5 steps, which provides an efficient trade-off between computation and trajectory quality while preserving multimodal behavior.

The flow-based action policy models the conditional distribution of action chunks using a normalizing flow with a standard Gaussian base distribution, enabling exact likelihood optimization. The flow is composed of multiple coupling layers with learned affine transformations, and each layer is conditioned on the same vision features and cached VLM hidden states as the diffusion model to ensure architectural parity. Actions are standardized to zero mean and unit variance, and training maximizes the conditional log-likelihood of expert demonstrations. At inference, action sequences are generated by sampling from the Gaussian base and applying the inverse flow with 5 flow steps; reducing the number of steps leads to noticeably degraded performance.

\textbf{Simulation Settings.} 
All experiments are conducted in a high-fidelity Isaac Sim simulator running at 30 Hz. The navigation policy operates at a control frequency of 10 Hz; actions are applied once every three simulation frames and held constant in between. Each episode is executed in an isolated process with a fresh simulator instance, ensuring clean initialization and reproducibility. Episodes terminate when the robot reaches the goal (defined by a 2D distance threshold of 1.5 m) or when a timeout is reached. Robot trajectories are recorded at the simulation rate to compute path length and SPL. Collisions are detected both with animated human characters using distance proximity (threshold of 0.5 m) and with static scene objects using a contact sensor. All metrics are aggregated across episodes and reported using standard navigation benchmarks.

\begin{algorithm}[ht]
\caption{Closed-loop Rollout with Asynchronous VLM Reasoning}
\label{alg:closed_loop_async}
\begin{algorithmic}[1]
\REQUIRE Environment $\mathcal{E}$ (30 Hz), wrapper TIC-VLA, controller $\mathcal{C}$, instruction $I$, goal $g$

\STATE $s_0 \gets \mathcal{E}.\textsc{Reset}()$
\STATE image buffer $\mathcal{B}\gets \emptyset$
\STATE reset episode state()
\STATE $\texttt{first\_kv\_ready}\gets \textbf{False}$
\STATE $(v_0,\omega_0)\gets(0,0)$ \COMMENT{initial controls}
\STATE $(v_k,\omega_k)\gets(v_0,\omega_0)$

\FOR{$k=0,1,2,\ldots$}
    \STATE $\texttt{pose}_k\gets \mathcal{E}.\textsc{ObservePose}()$
    \STATE $\texttt{img}_k\gets \mathcal{E}.\textsc{RenderCamera}()$
    \STATE append $\texttt{img}_k$ to $\mathcal{B}$

    \IF{$k \bmod 3 = 0$}
        \STATE $(\mathcal{X}^{\text{act}}_k,\mathcal{X}^{\text{vlm}}_k)\gets \textsc{SampleFramesAvailable}(\mathcal{B})$
        \STATE $s^{\text{robot}}_k\gets [v_{x,k},v_{y,k},\omega_{z,k}]$
        \COMMENT{robot state only; latency metadata is computed inside \texttt{predict\_async}}

        \STATE $(r_k,\mathbf{w}_k,k_{\text{gen}},\texttt{kv\_ready},\texttt{pose}_{\text{gen}})
        \gets \texttt{predict\_async}(\mathcal{X}^{\text{act}}_k,\mathcal{X}^{\text{vlm}}_k,I,s^{\text{robot}}_k,k,\texttt{pose}_k)$

        \IF{$\neg\,\texttt{first\_kv\_ready}\ \wedge\ \texttt{kv\_ready}$}
            \STATE $\texttt{first\_kv\_ready}\gets \textbf{True}$
        \ENDIF

        \IF{$\neg\,\texttt{first\_kv\_ready}$}
            \STATE $(v^{*}_k,\omega^{*}_k)\gets(0,0)$
        \ELSE
            \STATE $(v^{*}_k,\omega^{*}_k)\gets \mathcal{C}(\mathbf{w}_k)$
        \ENDIF

        \STATE $(v_k,\omega_k)\gets \textsc{Smooth}(v^{*}_k,\omega^{*}_k)$
    \ENDIF

    \STATE $\mathcal{E}.\textsc{Step}(v_k,\omega_k)$ \COMMENT{termination: success / timeout}
\ENDFOR
\end{algorithmic}
\end{algorithm}

\textbf{Testing Settings.}
We employ a closed-loop simulator running at 30\,Hz, while executing the policy at 10\,Hz (every 3 simulation steps), as summarized in \cref{alg:closed_loop_async}. At each 10~Hz tick, the agent samples a short visual context $\mathcal{X}_k$ from the image buffer using \textsc{SampleFramesAvailable}, which selects all available frames among the target offsets, e.g., $\{t-9,t-6,t-3,t\}$, and falls back to fewer frames when the full history is not yet available. Latency-aware metadata $(\Delta t,\Delta x,\Delta y,\Delta\theta)$ is computed inside \texttt{predict\_async} from the timestamp and pose associated with the cached VLM state. The controller $\mathcal{C}$ then consumes the policy output $\mathbf{w}_k$ to produce smoothed velocity commands $(v_k,\omega_k)$, and the environment advances with $\mathcal{E}.\textsc{Step}(v_k,\omega_k)$.

\textit{Asynchronous inference.}
VLM reasoning and KV-cache extraction are executed asynchronously via \texttt{predict\_async} (\cref{alg:predict_async_kv}), while the action policy continues to run at 10~Hz. Each call first polls the background VLM job and, if it has completed, updates the cached response $r^{\text{cache}}$, cached KV features $\texttt{kv}^{\text{cache}}$, and the timestamp and pose associated with the cached semantic state. If no generation is currently running, a new background generation job is launched using the VLM visual context $\mathcal{X}^{\text{vlm}}_k$ and instruction $I$. While the new VLM job is in flight, the action policy reuses the most recent available KV cache features to decode $\mathbf{w}_k$ from the current observation, robot state, and latency metadata.

\textit{Synchronous inference.}
In the synchronous baseline, VLM inference is executed in a blocking manner, during which the robot halts and no control commands are issued. At each decision step, the robot remains stationary while the VLM processes the current visual context and instruction, ensuring that semantic reasoning is generated from up-to-date observations. As a result, the effective semantic delay satisfies $\Delta t = 0$, and the corresponding ego-motion offsets $(\Delta x, \Delta y, \Delta \theta)$ are all zero. Once inference completes, the robot executes the predicted trajectory for a fixed horizon of 0.5 s, after which it again halts for the next VLM inference cycle. This design eliminates semantic staleness but introduces intermittent control stalls due to blocking inference.

\textit{Inference without reasoning.}
For ablations that remove explicit VLM reasoning, we keep the same closed-loop rollout (\cref{alg:closed_loop_async}) and the same asynchronous interface (\cref{alg:predict_async_kv}), but disable VLM generation entirely. Concretely, the background VLM job performs a forward pass on the visual-language input $\mathcal{X}^{\text{vlm}}_k$ and returns only the extracted KV cache; no tokens are decoded, and the textual response is not produced. As a result, $r_k$ is always \textbf{None}, and control depends only on the cached KV signal and current observations.

\textit{Latency control in simulation.}
We evaluate robustness under effective semantic delay, defined as the elapsed time between observation capture and semantic state availability. To isolate the effect of delay, we keep the semantic model fixed and vary the scheduling of semantic updates. The default asynchronous setting yields an average effective latency of about 2 s with full semantic reasoning. For a lower-latency condition (1 s), we pause the simulation after 1~s of elapsed time and resume it once VLM inference completes, yielding a reduced effective delay. For higher-latency settings (3-5 s), visual observations are captured immediately, but VLM inference is launched after a delay of 30, 60, or 90 simulation frames (30 Hz). During this period, the policy continues executing with the most recent cached semantics, and newly generated responses and KV caches are applied immediately upon completion. This increases effective semantic delay without altering simulation dynamics.

\begin{algorithm}[ht]
\caption{\texttt{predict\_async} (asynchronous VLM generation and action policy)}
\label{alg:predict_async_kv}
\begin{algorithmic}[1]
\STATE \textbf{Function} \texttt{predict\_async}$(\mathcal{X}^{\text{act}}_k,\mathcal{X}^{\text{vlm}}_k,I,s^{\text{robot}}_k,k,\texttt{pose}_k)$

\STATE \texttt{/* persistent internal state stored across calls */}
\STATE \hspace{1em} \texttt{gen\_job} \COMMENT{background VLM generation job, or \textbf{None}}
\STATE \hspace{1em} $\texttt{kv}^{\text{cache}}$ \COMMENT{most recent extracted KV cache}
\STATE \hspace{1em} $r^{\text{cache}}$ \COMMENT{most recent completed VLM response}
\STATE \hspace{1em} $k^{\text{cache}},\texttt{pose}^{\text{cache}}$ \COMMENT{step and pose associated with the cached KV}

\STATE \textsc{PollRunningGeneration}($k$)
\COMMENT{if a background generation finishes, update $\texttt{kv}^{\text{cache}},r^{\text{cache}},k^{\text{cache}},\texttt{pose}^{\text{cache}}$}

\STATE $k_{\text{gen}} \gets \textbf{None}$
\STATE $\texttt{pose}_{\text{gen}} \gets \textbf{None}$

\IF{\texttt{gen\_job} is inactive}
    \STATE $(k_{\text{gen}},\texttt{pose}_{\text{gen}})\gets(k,\texttt{pose}_k)$
    \COMMENT{record when and where this generation starts}
    \STATE start background VLM generation using $\mathcal{X}^{\text{vlm}}_k$, $I$, $k_{\text{gen}}$, and $\texttt{pose}_{\text{gen}}$
    \STATE \texttt{gen\_job} becomes active
\ENDIF

\IF{$\texttt{kv}^{\text{cache}}$ is undefined}
    \STATE $\texttt{kv}\gets \textbf{None}$
    \STATE $r_k\gets r^{\text{cache}}$
    \STATE $\texttt{kv\_ready}\gets \textbf{False}$
    \STATE $(\Delta t,\Delta x,\Delta y,\Delta\theta)\gets(0,0,0,0)$
\ELSE
    \STATE $\texttt{kv}\gets \texttt{kv}^{\text{cache}}$
    \STATE $r_k\gets r^{\text{cache}}$
    \STATE $\texttt{kv\_ready}\gets \textbf{True}$
    \STATE $\Delta t \gets \textsc{ElapsedTime}(k,k^{\text{cache}})$
    \STATE $(\Delta x,\Delta y,\Delta\theta)\gets \textsc{RelativePose}(\texttt{pose}^{\text{cache}},\texttt{pose}_k)$
\ENDIF

\STATE $\mathbf{w}_k\gets \pi_\theta(\textsc{Embed}(\mathcal{X}^{\text{act}}_k),\, s^{\text{robot}}_k,\, \texttt{kv},\,\Delta t,\Delta x,\Delta y,\Delta\theta)$
\COMMENT{use current vision, robot state, cached semantic KV, and latency metadata}

\STATE \textbf{return} $(r_k,\mathbf{w}_k,k_{\text{gen}},\texttt{kv\_ready},\texttt{pose}_{\text{gen}})$
\end{algorithmic}
\end{algorithm}

\textbf{Evaluation Metrics.}
We evaluate navigation performance using standard goal-reaching metrics. Success (S) for a single episode is defined as whether the robot reaches the goal within a 2D Euclidean distance threshold $T_{\text{goal}} = 1.5\text{m}$:
\begin{equation} 
\text{S} = \mathbb{I}\!\left( \lVert \mathbf{p}_{t,xy} - \mathbf{g}_{xy} \rVert_2 < T_{\text{goal}} \right), 
\end{equation}
where $\mathbf{p}_{t,xy}$ denotes the robot position at termination and $\mathbf{g}_{xy}$ denotes the goal position.
Success Rate (SR) is reported as the average success over all testing episodes.

Navigation Error (NE) measures the final 2D distance to the goal:
\begin{equation}
\text{NE} = \lVert \mathbf{p}_{t,xy} - \mathbf{g}_{xy} \rVert_2 .
\end{equation}

Path Length (PL) is computed as the cumulative 2D distance traveled by the robot:
\begin{equation}
\text{PL} = \sum_{i=1}^{N-1}
\lVert \mathbf{p}_{i+1,xy} - \mathbf{p}_{i,xy} \rVert_2 ,
\end{equation}
where $\{\mathbf{p}_{i,xy}\}_{i=1}^N$ are robot positions.

Success weighted by Path Length (SPL) is defined as:
\begin{equation}
\text{SPL} =
\text{SR} \cdot
\frac{d_{\text{opt}}}{\max(d_{\text{opt}}, \text{PL})},
\end{equation}
where $d_{\text{opt}}$ is the shortest distance from the start position $\mathbf{p}_{0,xy}$ to the goal.

Collision (C) is a binary episode-level indicator that equals 1 if any collision occurs during the episode and 0 otherwise:
\begin{equation} 
\text{C} = \mathbb{I}\!\left(\exists\, t \;\text{s.t.}\; \text{collision}_t \right), 
\end{equation}
Collisions include contacts with both animated human agents and static scene objects. The Collision Rate (CR) is computed as the fraction of test episodes in which at least one collision occurs.

\textbf{Physical Robot Testing.}  
We evaluate all methods on a Unitree Go2 quadruped robot. The primary computing setup consists of a laptop equipped with an NVIDIA RTX~4060 GPU and an NVIDIA Jetson Orin NX mounted onboard. For baseline methods whose memory or compute requirements exceed the capacity of these devices, we use a remote desktop equipped with an NVIDIA RTX~A6000 GPU. In this setup, the remote machine performs model inference and communicates with the laptop that relays sensor observations and action commands to the robot.

\section{Additional Results}

\begin{figure}[ht]
    \centering
    \includegraphics[width=0.98\linewidth]{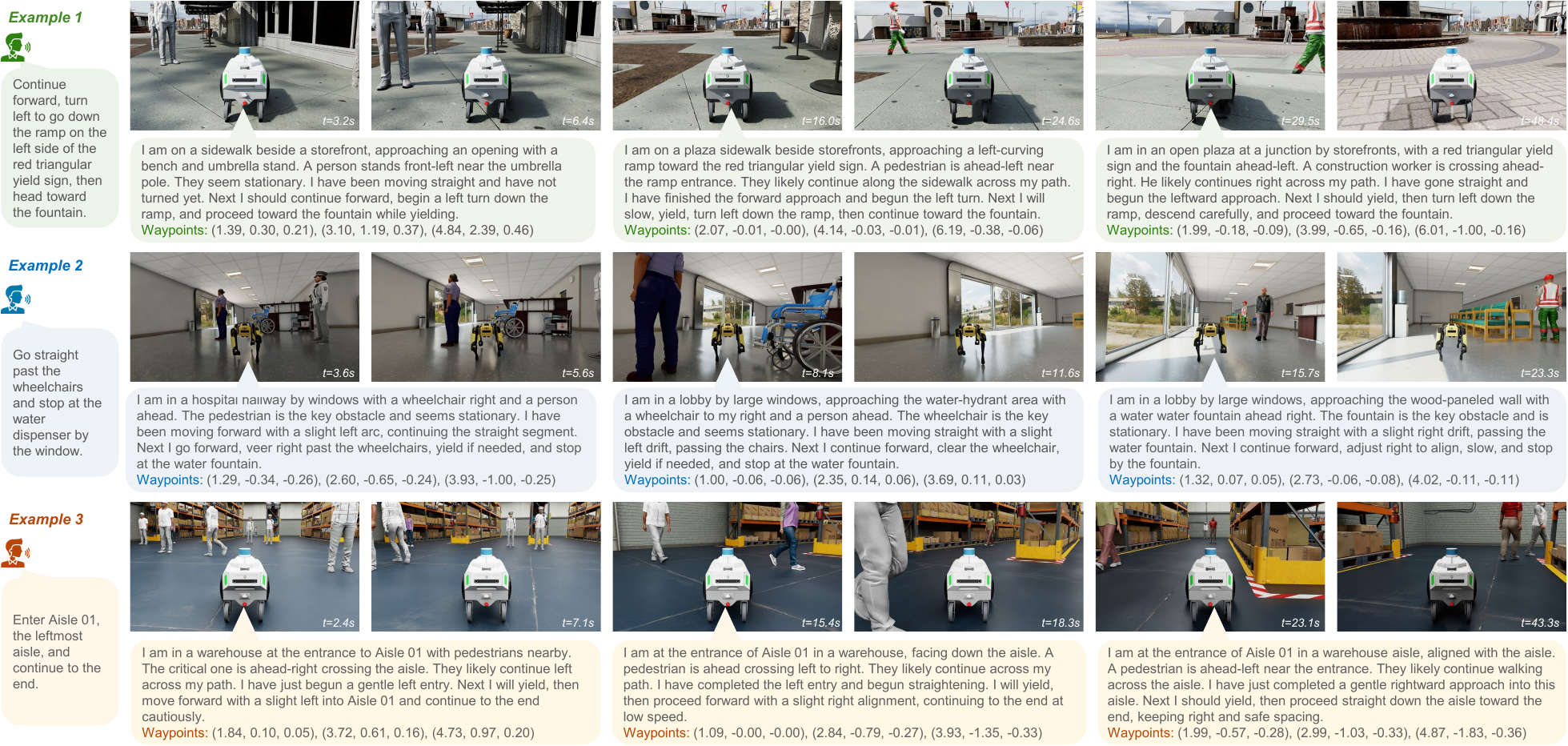}
    \caption{Additional qualitative simulation results of TIC-VLA on the DynaNav benchmark. Rows correspond to different environment types: \emph{top}-outdoor, \emph{middle}-hospital, and \emph{bottom}-warehouse.}
    \label{fig:sim}
\end{figure}

\textbf{Simulation Testing.}
We provide additional simulation testing results of TIC-VLA in diverse DynaNav environments, including outdoor, hospital, and warehouse, using multiple robot platforms, in \cref{fig:sim}. The results demonstrate that TIC-VLA is able to generate coherent semantic reasoning under delayed perception and translate it into stable, low-level control commands. The robot consistently adapts its behavior to dynamic agents by yielding, adjusting speed, and selecting appropriate turning maneuvers, while following the given natural language instructions. To further illustrate temporal consistency and closed-loop behavior, dynamic video results are provided. 

\begin{figure}[ht]
    \centering
    \includegraphics[width=0.95\linewidth]{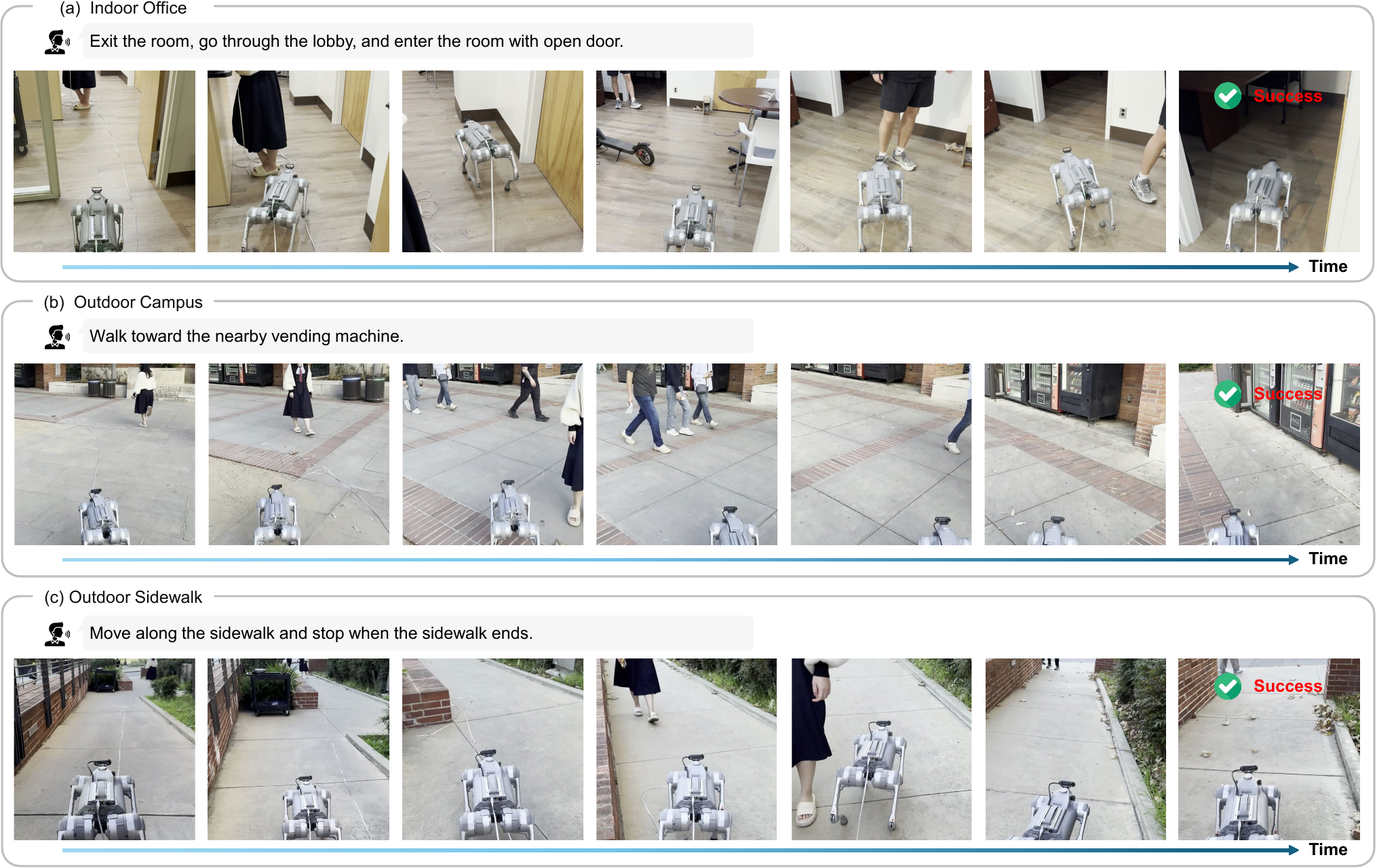}
    \caption{Real-world navigation examples demonstrating \textbf{TIC-VLA} executing language-conditioned navigation under real-time control.}
    \label{fig:real_test}
\end{figure}

\textbf{Real-world Testing.}
\textbf{Real-world Testing.}
We report task-specific results from real-world robot navigation experiments. Each task is executed five times, and the average success rate (SR) is summarized in \cref{tab:real}. As shown, TIC-VLA consistently outperforms prior methods across both indoor and outdoor environments. \cref{fig:real_test} presents three representative real-world navigation examples under diverse conditions. To further demonstrate the closed-loop control behavior, we provide additional dynamic video results on the \href{https://ucla-mobility.github.io/TIC-VLA/#results}{project website}.

\begin{table}[ht]
\centering
\small
\caption{Real-world testing task-specific evaluation results.}
\label{tab:real}
\begin{tabular}{l | c | c | c}
\toprule
\textbf{Task} 
& \textbf{TIC-VLA SR (\%)} 
& \textbf{DualVLN SR (\%)} 
& \textbf{NaVILA  SR (\%)} \\
\midrule
Indoor Hallway      & 80  & 40 & 20 \\
Indoor Office       & 80  & 20 & 0 \\
Outdoor Campus      & 80  & 60 & 60 \\
Outdoor Sidewalk    & 100 & 80 & 60 \\ 
\midrule
Average             & 85 & 50 & 35 \\
\bottomrule
\end{tabular}
\end{table}

\noindent\textbf{Influence of Action Policy.}
\Cref{tab:policy_ablation} compares three action policy designs in TIC-VLA: diffusion-based, flow-based, and query-based Transformer policies. Overall, we observe no performance benefit from the more complex diffusion and flow-matching policies over a simple query-based Transformer.
The query-based policy consistently achieves the best results across all metrics. By directly predicting action sequences in a single forward pass, it minimizes inference latency and preserves tight alignment between perception and control. Moreover, the simpler architecture facilitates more stable and efficient RL fine-tuning.

\begin{table}[ht]
\centering
\caption{Comparison of different action policies in the TIC-VLA model.}
\renewcommand{\arraystretch}{1.0}
\begin{tabular}{l|cccc}
\toprule
\textbf{Policy Type} & \textbf{NE ($\downarrow$)} & \textbf{SR ($\uparrow$)} & \textbf{SPL ($\uparrow$)} & \textbf{CR ($\downarrow$)} \\
\midrule
Diffusion-based     & 14.30 & 38.82 & 35.41 & 41.18 \\
Flow-based          & 15.04 & 32.94 & 29.65 & 45.88 \\
Query-based (Ours)  & \textbf{10.85}  & \textbf{47.06}& \textbf{42.41} & \textbf{34.12} \\
\bottomrule
\end{tabular}
\label{tab:policy_ablation}
\end{table}

\textbf{Influence of VLM Backbone.} 
We study the effect of VLM backbone scale and visual tokenization in TIC-VLA by comparing InternVL3 (1B) with SmolVLM2~\cite{shukor2025smolvla} (500M) and Qwen2.5-VL~\cite{bai2025qwen2} (3B), as summarized in \Cref{tab:backbone_ablation}. The smaller SmolVLM2 benefits from lower inference latency but exhibits higher navigation error and lower success and SPL due to limited representational capacity. In contrast, the larger Qwen2.5-VL provides stronger visual understanding but incurs substantial inference delay, which degrades perception-action alignment and overall navigation performance. InternVL3 achieves the best trade-off between capacity and efficiency.

\begin{table}[ht]
\centering
\caption{Comparison of VLM backbone in the TIC-VLA model. }
\label{tab:backbone_ablation}
\begin{tabular}{l|cccc}
\toprule
\textbf{Backbone} & \textbf{NE ($\downarrow$)} & \textbf{SR ($\uparrow$)} & \textbf{SPL ($\uparrow$)} & \textbf{CR ($\downarrow$)} \\
\midrule
SmolVLM2 (500M) & 14.82 & 38.82 & 31.06 & 41.18 \\
InternVL3 (1B)  & \textbf{10.85}  & \textbf{47.06}& \textbf{42.41} & \textbf{34.12} \\
Qwen2.5-VL (3B) & 13.26 & 32.94 & 29.98 & 35.29 \\
\bottomrule
\end{tabular}
\end{table}

\textbf{Influence of Action Expert Architecture.}
We evaluate the sensitivity of TIC-VLA to the depth of the Transformer action expert by varying the number of cross-attention layers. All variants are evaluated without RL fine-tuning.
As shown in \cref{tab:architecture_ablation}, the 6-layer design achieves the strongest overall performance. The 3-layer variant performs comparably but is slightly weaker across most metrics, suggesting that a shallower action expert has limited capacity to integrate visual tokens, semantic features, and latency metadata. Increasing the depth to 12 layers does not improve performance and leads to a worse success rate, SPL, and collision rate despite a slightly lower navigation error. We therefore adopt the 6-layer action expert as the default architecture.

\begin{table}[ht]
\centering
\caption{Effect of the number of cross-attention layers in the Transformer action expert. }
\renewcommand{\arraystretch}{1.05}
\begin{tabular}{l|cccc}
\toprule
\textbf{Layers} & \textbf{NE ($\downarrow$)} & \textbf{SR ($\uparrow$)} & \textbf{SPL ($\uparrow$)} & \textbf{CR ($\downarrow$)} \\
\midrule
3 & 10.98 & 45.88 & 41.39 & 35.29 \\
6 (ours) & 10.85 & \textbf{47.06} & \textbf{42.41} & \textbf{34.12} \\
12 & \textbf{10.73} & 44.71 & 40.77 & 37.64 \\
\bottomrule
\end{tabular}
\label{tab:architecture_ablation}
\end{table}

\textbf{Sensitivity to Odometry Drift.} 
We further evaluate robustness by injecting Gaussian noise into translation and rotation measurements during simulation. The noise magnitude is proportional to the traveled distance at each step and accumulates over time, simulating realistic odometry drift.
As shown in \cref{tab:drift_noise}, TIC-VLA remains stable under typical drift conditions, with only marginal changes in navigation error, success rate, and SPL. Even under the severe stress-test setting, the policy continues to navigate successfully in a substantial portion of episodes, although performance degradation becomes more apparent due to accumulated localization errors. These results suggest that TIC-VLA is reasonably robust to moderate odometry inaccuracies and does not catastrophically fail under noisy motion estimation.

\begin{table}[ht]
\centering
\caption{Robustness of TIC-VLA under accumulated localization drift with Gaussian translation and rotation noise.}
\label{tab:drift_noise}
\setlength{\tabcolsep}{4pt}
\renewcommand{\arraystretch}{1.0}
\begin{tabular}{l|c|cccc}
\toprule
\textbf{Drift Level} & \textbf{Gaussian Noise (trans, rot)} & \textbf{NE ($\downarrow$)} & \textbf{SR ($\uparrow$)} & \textbf{SPL ($\uparrow$)} & \textbf{CR ($\downarrow$)} \\
\midrule
None    & --                    & 10.55 & 55.29 & 50.29 & 28.24 \\
Typical & (0.02 m/m, $1^\circ$/m) & 10.66 & 55.29 & 50.12 & 29.41 \\
Severe  & (0.05 m/m, $5^\circ$/m) & 11.87 & 43.53 & 38.41 & 35.29 \\
\bottomrule
\end{tabular}
\end{table}




\end{document}